\documentclass[journal]{IEEEtran}
\usepackage{amsmath,graphicx}
\usepackage[T1]{fontenc}
\usepackage{hyperref, times, cite}
\usepackage{url, datetime}
\usepackage{mathtools}
\usepackage{bm, bbm}
\usepackage{cool}
\usepackage{amssymb,amsfonts,amsmath,amsthm,amscd,dsfont,mathrsfs}
\usepackage{graphicx,float,psfrag,epsfig,amssymb}
\usepackage{wrapfig}
\usepackage{relsize}
\usepackage{color}
\usepackage{pict2e}
\usepackage{algorithm}
\usepackage{algorithmic}
\usepackage{caption}
\usepackage{nameref}
\usepackage{enumerate}
\usepackage{stackrel}
\usepackage{enumitem}
\usepackage{caption}
\usepackage{subcaption}

\usepackage{amsthm}

\begin{document}
\thispagestyle{empty}

\title{Learning Sparse High-Dimensional Matrix-Valued Graphical Models From Dependent Data}
\author{ Jitendra K.\ Tugnait \vspace*{-0.3in}
\thanks{J.K.\ Tugnait is with the Dept. of 
Elec.\ \& Comp.\ Eng.,
200 Broun Hall, Auburn University, Auburn, AL 36849, USA. 
Email: tugnajk@auburn.edu . }

\thanks{This work was supported by the National Science Foundation Grants ECCS-2040536 and CCF-2308473.}}

\maketitle

\renewcommand{\algorithmicrequire}{\textbf{Input:}}
\renewcommand{\algorithmicensure}{\textbf{Output:}}

\begin{abstract}
We consider the problem of inferring the conditional independence graph (CIG) of a sparse, high-dimensional, stationary matrix-variate Gaussian time series. All past work on high-dimensional matrix graphical models assumes that independent and identically distributed (i.i.d.) observations of the matrix-variate are available. Here we allow dependent observations. We consider a sparse-group lasso-based frequency-domain formulation of the problem with a Kronecker-decomposable power spectral density (PSD), and solve it via an alternating direction method of multipliers (ADMM) approach. The problem is bi-convex which is solved via flip-flop optimization. We provide sufficient conditions for local convergence in the Frobenius norm of the inverse PSD estimators to the true value. This result also yields a rate of convergence. We illustrate our approach using numerical examples utilizing both synthetic and real data.  
\end{abstract}

\vspace*{-0.15in}
\section{Introduction} \label{intro}

\IEEEPARstart{I}{n} graphical models, graphs display the conditional independence structure of the variables, and learning the graph structure is equivalent to learning a factorization of the joint probability distribution of these random variables \cite{Lauritzen1996}. In a vector graphical model, the conditional statistical dependency structure among $p$ random variables $x_1, x_2, \cdots , x_p$, is represented using an undirected graph ${\cal G} = \left( V, {\cal E} \right)$ with a set of $p$ vertices (nodes) $V = \{1,2, \cdots , p\} =[p]$, and a corresponding set of (undirected) edges ${\cal E} \subseteq [p] \times [p]$. There is no edge between nodes $i$ and $j$ iff $x_i$ and $x_j$ are conditionally independent given the remaining $p$-$2$ variables. Suppose ${\bm x} \sim {\mathcal N}_r( {\bf m}, {\bm \Sigma})$, with ${\bf m} \in \mathbb{R}^p$, ${\bm \Sigma} \in \mathbb{R}^{p \times p}$, positive definite ${\bm \Sigma} = {\bm \Omega}^{-1}$, where ${\mathcal N}_r( {\bf m}, {\bm \Sigma})$ denotes a real-valued Gaussian vector with mean ${\bm m}$ and covariance ${\bm \Sigma}$. Then ${\Omega}_{ij}$, the $(i,j)$-th element of  $\bm{\Omega}$, is zero iff $x_i$ and $x_j$ are conditionally independent \cite{Lauritzen1996}. Of much interest is the high-dimensional case where $p$ is greater than or of the order of the data sample size $n$ \cite{Wainwright2019}. In particular, in a high-dimensional setting, as $n \uparrow \infty$, $p/n \rightarrow c > 0$, instead of $p/n \rightarrow 0$ as in classical low-dimensional statistical analysis framework \cite[Chapter 1]{Wainwright2019}. Such models for ${\bm x}$ have been extensively studied \cite{Wainwright2019, Meinshausen2006, Banerjee2008, Friedman2008}. In this paper we address the problem of high-dimensional matrix graph estimation. If $p/n \ll 1$, we use the term low-dimensional for such cases in this paper.

Consider a stationary $p-$dimensional multivariate Gaussian time series ${\bm x}(t)$, $t=0, \pm 1, \pm 2, \cdots $, with $i$th component $x_i(t)$.  In the corresponding time series graph ${\cal G} = \left( V, {\cal E} \right)$, there is no edge between nodes $i$ and $j$ iff $\{ x_i(t) \}$ and $\{ x_j(t) \}$ are conditionally independent given the remaining $p$-$2$ scalar series $\{ x_\ell(t), \, \ell \in [p], \; \ell \neq i, \, \ell \neq j \}$ \cite{Dahlhaus2000}. 
Denote the power spectral density (PSD) matrix of zero-mean $\{ {\bm x}(t) \}$ by ${\bm S}_x(f)$, where ${\bm S}_x(f) = \sum_{\tau = -\infty}^{\infty}  {\bm R}_{xx}( \tau ) e^{-\iota 2 \pi f \tau}$, ${\bm R}_{xx}( \tau ) = E\{ {\bm x}(t+\tau) {\bm x}^\top (t) \}$ and $\iota = \sqrt{-1}$. In \cite{Dahlhaus2000} it was shown that conditional independence of two time series components given all other components of the zero-mean time series, is encoded by zeros in the inverse PSD, that is, $\{ i,j \} \not\in {\cal E}$ iff the $(i,j)$-th element of ${\bm S}_x^{-1}(f)$,  $[{\bm S}_x^{-1}(f)]_{ij} = 0$ for every $f$. In \cite{Dahlhaus2000} the low-dimensional case is addressed whereas nonparametric frequency-domain approaches for graph estimation in high-dimensional settings have been considered in \cite{Jung2015a, Tugnait18c, Tugnait22c}. Refs.\ \cite{Jung2015a, Tugnait22c} provide performance analysis and guarantees. Parametric modeling based approaches in low-dimensional settings for conditional independence graph (CIG) estimation for time series  are discussed in \cite{Avventi2013, Zorzi2016, Alpago2018, Songsiri2010, Ciccone2020, Alpago2023}. These papers are focused on algorithm development and they do not provide performance guarantees (such as \cite[Theorem 1]{Tugnait22c}). Estimation of sparse high-dimensional parametric time series models is discussed in \cite{Basu15} where performance analysis in high-dimensions is carried out, but the graphical modeling aspect is not addressed.
\vspace*{-0.01in}

The need for matrix-valued graphical models arises in several applications \cite{Leng2012, Yin2012, Tsiligkaridis2013, He2014, Zhou2014, Huang2015, Zhu2018, Chen2019, Greenewald2019, Lyu2020, Min2021} (see also related work of \cite{Werner2008}). Here we observe matrix-valued time series $\{ {\bm Z}(t)\}$ where ${\bm Z}(t) \in \mathbb{R}^{p \times q}$. If one vectorizes using $\mbox{vec}({\bm Z})$ where $\mbox{vec}({\bm Z}) \in \mathbb{R}^{pq}$ denotes column-wise vectorization of ${\bm Z}$, then use of $\mbox{vec}({\bm Z})$ will result in a $pq$-node graph with $(pq) \times (pq)$ precision matrix, which could be ultra-high-dimensional, and it ignores any structural information among rows and columns of ${\bm Z}(t)$ \cite{Leng2012}. With $\otimes$ denoting the matrix Kronecker product, the basic idea in matrix-valued graphs is to model the covariance of $\mbox{vec}({\bm Z})$ as ${\bm \Psi} \otimes {\bm \Sigma}$ with ${\bm \Psi} \in \mathbb{R}^{q \times q}$ and ${\bm \Sigma} \in \mathbb{R}^{p \times p}$, reducing the number of unknowns from ${\cal O}(p^2 q^2)$  to ${\cal O}(p^2 + q^2)$, while also preserving the structural information. Given data, one estimates two precision matrices ${\bm \Omega} = {\bm \Sigma}^{-1}$ and ${\bm \Upsilon} = {\bm \Psi}^{-1}$. In the matrix graph, conditional independence between ${\bm Z}_{ij}$ and ${\bm Z}_{k \ell}$ is determined by zeros in ${\bm \Omega}$ and ${\bm \Upsilon}$ \cite{Leng2012}. This is the Kronecker graph model 
\cite{Weichsel1962, Leskovec2010}: If ${\cal G}_1$ and ${\cal G}_2$ are graphs with adjacency matrices ${\cal A}({\cal G}_1)$ and ${\cal A}({\cal G}_2)$, respectively, then the Kronecker product graph (KPG) ${\cal G}_1 \otimes {\cal G}_2$ is defined as the graph with adjacency matrix ${\cal A}({\cal G}_1) \otimes {\cal A}({\cal G}_2)$ \cite[Def.\ 1]{Leskovec2010}. In our context the nonzero entries of ${\bm \Upsilon}$ and ${\bm \Omega}$ determine the nonzero entries of the adjacency matrices of graphs ${\cal G}_1$ and ${\cal G}_2$, respectively, with KPG ${\cal G} = {\cal G}_1 \otimes {\cal G}_2$. 

Our objective in this paper is to learn a conditional independence KPG associated with time-dependent matrix-valued zero-mean $p \times q$ Gaussian sequence ${\bm Z}(t)$,  under high-dimensional settings, given observations of $\{ {\bm Z}(t) \}_{t=0}^{n-1}$.
\vspace*{-0.12in}

\subsection{Related Work}
\vspace*{-0.05in} 
Prior work on KPG estimation under high-dimensional settings \cite{Leng2012, Yin2012, Tsiligkaridis2013, He2014, Zhou2014, Huang2015, Zhu2018, Chen2019, Greenewald2019, Lyu2020, Min2021} all assume that i.i.d.\ observations of ${\bm Z}$ are available for graphical modeling. Refs.\ \cite{Leng2012, Tsiligkaridis2013, Yin2012} all solve the same bi-convex optimization problem, using an identical alternating minimization approach, but they differ in theoretical analysis. Ref.\  \cite{Greenewald2019} uses a Kronecker sum model whereas we use a Kronecker-product separable covariance structure (see (\ref{eqn125}) later) for $\{ {\bm Z}(t) \}$. 

There is no prior reported work on high-dimensional matrix graph estimation with dependent data using a nonparametric approach. Parametric models using state-space models are estimated in \cite{Carvalho2007, Wang2009} for KPG estimation in low-dimensional settings using Bayesian approaches. Granger causality graphs (not the same as CIGs) for matrix time series are estimated in \cite{Jiang2021} using first-order AR models and in \cite{Zorzi2022} using an infinite dimensional model class (which includes ARMAX models of any order), both in low-dimensional settings (i.e., $pq/n \ll 1$ and/or $\lim_{n \rightarrow \infty} pq/n =  0$, with $pq$ representing number of nodes in KPG). In contrast, this paper considers conditional independence KPG's under high-dimensional settings. Ref.\ \cite{Sinquin2019} investigates sum of Kronecker product AR models for matrix times series with no consideration of CIGs. Estimation of a KPG model corresponding to an AR Gaussian process is investigated in \cite{Zorzi2020} in low-dimensional settings with no performance analysis or guarantees. A distinguishing aspect of \cite{Zorzi2020} is that it imposes a Kronecker product decomposition on the support of the inverse PSD, not the inverse PSD of the time series. With regard to \cite{Zorzi2022, Zorzi2020}, we note that in the synthetic data example using an ARMA model in \cite[Sec.\ 6.1, Fig.\ 2]{Zorzi2022}, number of nodes is 16 $(=pq)$ and sample size is $n=3900$, leading to $pq/n = 0.004$, a low-dimensional setting. The real data example of \cite[Sec.\ 6.2]{Zorzi2022} does have $pq=$96 and $n=500$, implying $pq/n = 0.19$. The distinction is that the ground truth is known in synthetic data examples permitting evaluation of the efficacy of the considered approach, whereas such is not the case in real data examples. Thus \cite{Zorzi2022} does not address the high-dimensional scenario as relatively high $pq/n=0.19$ in their real data example is not supported by any commensurate synthetic data example. In contrast, we provide such support, as seen in Table I, Sec.\ VI-A of this paper, where $pq=$225 and varying $n \in \{64, 128, 256, 512, 1024, 2148\}$, implying $pq/n = \{ 3.5, 1.76, 0.88, 0.44, 0.22, 0.11\}$. In our real data example (Sec.\ VI-B), we have $pq=88$ and $n=364$ with $pq/n = 0.24$. In \cite[Sec.\ 6.1]{Zorzi2020}, the synthetic data example has $pq=36$ and $n=$1000 or 2000 ($pq/n$=0.036 or 0.018), again a low-dimensional scenario. The real data example of \cite[Sec.\ 6.2]{Zorzi2020} has $pq=36$ and $n=389$ ($pq/n = 0.09$). The comments made pertaining to \cite{Zorzi2022} regarding differences in $pq/n$ ratios for real and synthetic data examples, apply to \cite{Zorzi2020} as well.
Finally, \cite{Chen2021} considers a first-order matrix AR model for matrix time series where when vectorized, the vectorized time-series AR coefficient is expressed as a Kronecker product. Low-dimensional asymptotics are provided in \cite{Chen2021} and the issue of the underlying CIG is not addressed. 

A frequency-domain formulation is used in this paper, following the approach of \cite{Tugnait22c} for dependent vector time series. The resulting optimization problem is bi-convex, as in \cite{Leng2012, Yin2012, Tsiligkaridis2013}, but with complex variables, and is solved via an alternating minimization approach using Wirtinger calculus \cite{Schreier10} for optimization of real functions of complex variables.  

A preliminary version of parts of this paper appear in a workshop paper \cite{Tugnait23}. Theorems 1-3 and their proofs, and the real data example do not appear in \cite{Tugnait23}. 

\vspace*{-0.2in}
\subsection{Our Contributions, Outline and Notation} 
\vspace*{-0.05in}
The underlying system model including a generative model (\ref{eqn128}) for time-dependent matrix Gaussian sequence, is presented in Sec.\ \ref{SM}.  A frequency-domain based penalized log-likelihood objective function is derived in Sec.\ \ref{pnll} for estimation of the matrix graph, resulting in a Kronecker-decomposable power spectral density representation (\ref{eqn213}).  A flip-flop algorithm based on two ADMM algorithms is presented in Sec.\ \ref{opt} to optimize the bi-convex objective function. In Sec.\ \ref{consist} the performance of the proposed optimization algorithm is analyzed under a high-dimensional large sample setting in Theorems 1-3, patterned after \cite{Lyu2020} and exploiting some results from \cite{Tugnait22c, Tugnait21, Rothman2008}. Numerical results are presented in Sec.\ \ref{NE} and proofs of Theorems 1, 2 and 3 are given in three appendices. 

{\it Notation}. The superscripts $\ast$, $\top$ and $H$ denote the complex conjugate, transpose and  conjugate transpose operations, respectively, $\mathbb{R}$ and $\mathbb{C}$ denote the sets of real and complex numbers, respectively, and $\mbox{Re}({\bm x})$ is the real part of ${\bm x} \in \mathbb{C}^p$. We use $\iota := \sqrt{-1}$. A $p \times p$ identity matrix is denoted by ${\bm I}_p$.  Given  ${\bm A} \in \mathbb{C}^{p \times p}$,  $\phi_{\min }({\bm A})$, $\phi_{\max }({\bm A})$, $|{\bf A}|$, $\mbox{tr}({\bm A})$ and $\mbox{etr}({\bm A})$ denote the minimum eigenvalue, maximum eigenvalue, determinant, trace, and exponential of trace of ${\bm A}$, respectively. We use ${\bm A} \succeq 0$ and ${\bm A} \succ 0$ to denote that Hermitian ${\bm A}$ is positive semi-definite and positive definite, respectively. 
For ${\bm B} \in \mathbb{C}^{p \times q}$, we define  the operator norm, the Frobenius norm and the vectorized $\ell_1$ norm, respectively, as $\|{\bm B}\| = \sqrt{\phi_{\max }({\bm B}^H  {\bm B})}$, $\|{\bm B}\|_F = \sqrt{\mbox{tr}({\bm B}^H  {\bm B})}$ and $\|{\bm B}\|_1 = \sum_{i,j} |B_{ij}|$, where $B_{ij}$ is the $(i,j)$-th element of ${\bm B}$, also denoted by $[{\bm B}]_{ij}$. 
For vector ${\bm \theta} \in \mathbb{C}^p$, we define $\| {\bm \theta} \|_1 = \sum_{i=1}^p |\theta_i|$ and $\| {\bm \theta} \|_2 = \sqrt{\sum_{i=1}^p |\theta_i|^2}$, and we also use $\| {\bm \theta} \|$ for $\| {\bm \theta} \|_2$. Given ${\bm A} \in \mathbb{C}^{p \times p}$, ${\bm A}^+ = \mbox{diag}({\bm A})$ is a diagonal matrix with the same diagonal as ${\bm A}$, and  ${\bm A}^- = {\bm A} - {\bm A}^+$ is ${\bm A}$ with all its diagonal elements set to zero. We use ${\bm A}^{-\ast}$ for $({\bm A}^\ast)^{-1}$, the inverse of complex conjugate of ${\bm A}$, and ${\bm A}^{- \top}$ for $({\bm A}^\top)^{-1}$. Given ${\bm A} \in \mathbb{C}^{n \times p}$, column vector $\mbox{vec}({\bm A}) \in \mathbb{C}^{np}$ denotes the vectorization of ${\bm A}$ which stacks the columns of the matrix ${\bm A}$. The notation ${\bm y}_n = {\cal O}_P({\bm x}_n)$ for random ${\bm y}_n, {\bm x}_n \in \mathbb{C}^p$ means that for any $\varepsilon > 0$, there exists $0 < T < \infty$ such that $P ( \|{\bm y}_n\| \le T \|{\bm x}_n\|) \ge 1 - \varepsilon$ $\forall n \ge 1$. The notation ${\bm x} \sim {\mathcal N}_c( {\bf m}, {\bm \Sigma})$ denotes a complex random vector  ${\bm x}$ that is circularly symmetric (proper), complex Gaussian with mean ${\bm m}$ and covariance ${\bm \Sigma}$, and ${\bm x} \sim {\mathcal N}_r( {\bf m}, {\bm \Sigma})$ denotes real-valued Gaussian ${\bm x}$ with mean ${\bm m}$ and covariance ${\bm \Sigma}$. 
\vspace*{-0.14in}

\section{System Model} \label{SM}
A random matrix ${\bm Z} \in \mathbb{R}^{p \times q}$ is said to have a matrix normal (Gaussian) distribution  if its pdf $f({\bm Z} | {\bm M}, {\bm \Sigma}, {\bm \Psi})$, characterized by ${\bm M} \in \mathbb{R}^{p \times q}$, ${\bm \Sigma} \in \mathbb{R}^{p \times p}$, ${\bm \Psi} \in \mathbb{R}^{q \times q}$, is \cite[Chap.\ 2]{Gupta1999}
\begin{equation} \label{eqn100}
  f({\bm Z} | {\bm M}, {\bm \Sigma}, {\bm \Psi}) = 
	  \frac{\mbox{etr} \Big( -\frac{1}{2}({\bm Z}-{\bm M}) {\bm \Psi}^{-1} 
		 ({\bm Z}-{\bm M})^\top {\bm \Sigma}^{-1} \Big)}{(2 \pi)^{pq/2} \, |{\bm \Sigma}|^{q/2} \, |{\bm \Psi}|^{p/2}} \, .
\end{equation}
We will use the notation ${\bm Z} \sim \mathcal{MVN}({\bm M}, {\bm \Sigma}, {\bm \Psi})$ for the matrix normal distribution specified by (\ref{eqn100}). Equivalently, 
\begin{equation} \label{eqn105}
  \mbox{vec}({\bm Z}) \, \sim \, {\cal N}_r \big( \mbox{vec}({\bm M}), {\bm \Psi} \otimes {\bm \Sigma} \big) \, .
\end{equation}
Here ${\bm \Psi}$ is the row covariance matrix and ${\bm \Sigma}$ is the column covariance matrix \cite{Gupta1999} since the $k$th column ${\bm Z}_{ \cdot k} \sim {\cal N}_r({\bm 0}, [{\bm \Psi}]_{kk} {\bm \Sigma})$ and the $i$th row ${\bm Z}_{i \cdot}^\top \sim {\cal N}_r({\bm 0}, [{\bm \Sigma}]_{ii} {\bm \Psi})$.

With ${\bm Z} \in \mathbb{R}^{p \times q}$ modeled as a zero-mean matrix normal vector and ${\bm z} = \mbox{vec}({\bm Z})$, \cite{Leng2012} assumes
\begin{align} 
  E\{ {\bm z} {\bm z}^\top \} = & {\bm \Psi} \otimes {\bm \Sigma} \, ,
		        \label{eqn109}
\end{align}
implying a separable covariance structure \cite{Werner2008}. Let ${\bm \Omega} = {\bm \Sigma}^{-1}$ and ${\bm \Upsilon} = {\bm \Psi}^{-1}$ denote the respective precision matrices. Then ${\bm Z}_{ij}$ and ${\bm Z}_{k \ell}$ are conditionally independent given remaining entries in ${\bm Z}$ iff (i) at least one of ${\bm \Omega}_{ik}$ and ${\bm \Upsilon}_{j \ell}$ is zero when $i\ne k$, $j \ne \ell$, (ii) ${\bm \Omega}_{ik}=0$ when $i\ne k$, $j = \ell$, and (iii) ${\bm \Upsilon}_{j \ell}=0$ when $i = k$, $j \ne \ell$ \cite{Leng2012}. 

In this paper we will model our time-dependent zero-mean matrix-valued, stationary, $p \times q$ Gaussian sequence ${\bm Z}(t)$, ${\bm z}(t) = \mbox{vec}({\bm Z}(t))$, as having the separable covariance structure given by
\begin{align} 
  E\{ {\bm z}(t+\tau) {\bm z}^\top(t) \} 
		= & {\bm \Psi}(\tau) \otimes {\bm \Sigma} \label{eqn125}
\end{align}
where ${\bm \Psi}(\tau)$, $\tau = 0, \pm 1, \cdots$ models time-dependence while ${\bm \Sigma} \succ {\bm 0}$ is fixed. Under (\ref{eqn125}), the row covariance sequence is $E \{ {\bm Z}_{i \cdot}^\top(t+\tau) {\bm Z}_{i \cdot}(t) \} = [{\bm \Sigma}]_{ii} \, {\bm \Psi}(\tau)$ and the column covariance sequence is $E \{ {\bm Z}_{ \cdot k}(t+\tau) {\bm Z}_{ \cdot k}^\top(t) \} = {\bm \Sigma} \, [{\bm \Psi}(\tau)]_{kk}$. Thus we allow possible temporal dependence in matrix observations via ${\bm \Psi}(\tau)$. 
With $\{ {\bm e}(t) \}$ i.i.d., ${\bm e}(t) \sim {\cal N}_r({\bm 0}, {\bm I}_{pq})$, a generative model for ${\bm z}(t)$ is given by
\begin{align} 
 & {\bm z}(t) =  \sum_{i=0}^L  ({\bm B}_i \otimes {\bm F}) {\bm e}(t-i) \, , \; {\bm B}_i \in {\mathbb R}^{q \times q}
	 \, , \; {\bm F} \in {\mathbb R}^{p \times p} \label{eqn128} \\
&	\Rightarrow \;  E\{ {\bm z}(t+\tau) {\bm z}^\top(t) \} 
		=  \big (\underbrace{  \sum_{i=0}^L {\bm B}_i {\bm B}_{i-\tau}^\top  }_{ = {\bm \Psi}(\tau) } \big)
		     \otimes \underbrace{ ( {\bm F} {\bm F}^\top ) }_{ = {\bm \Sigma} } \, . \label{eqn129}
\end{align}
In (\ref{eqn128}), we can have $L \uparrow \infty$ so long as assumption (A2) stated in Sec.\ \ref{pnll} holds. In sequel, we exploit (\ref{eqn125}) in our approach without considering (\ref{eqn129}), the latter is used only for synthetic data generation.

The PSD of $\{ {\bm z}(t) \}$ is ${\bm S}_z(f) = \bar{\bm S}(f) \otimes {\bm \Sigma}$ where $\bar{\bm S}(f) = \sum_{\tau} {\bm \Psi}(\tau) e^{-\iota 2 \pi f \tau}$. Then ${\bm S}_z^{-1}(f) = \bar{\bm S}^{-1}(f) \otimes {\bm \Sigma}^{-1}$, and by \cite{Dahlhaus2000}, in the $pq-$node graph ${\cal G} = \left( V, {\cal E} \right)$, $|V|=pq$,  associated with $\{ {\bm z}(t) \}$, edge $\{i,j\} \not\in {\cal E}$ iff $[{\bm S}_z^{-1}(f)]_{ij} = 0$ for every $f$. This does not account for the separable structure of our model. Noting that $\bar{\bm S}^{-1}(f)$, $f \in [0,0.5]$, plays the role of ${\bm \Upsilon} = {\bm \Psi}^{-1}$, using \cite{Leng2012, Dahlhaus2000} (also \cite[Observation 1]{Leskovec2010}), we deduce that $\{ {\bm Z}_{ij}(t) \}$ and $\{ {\bm Z}_{k \ell}(t) \}$ are conditionally independent given remaining entries in $\{ {\bm Z}(t) \}$ iff (i) at least one of ${\bm \Omega}_{ik}$ and $[\bar{\bm S}^{-1}(f)]_{j \ell}$, for every $f \in [0,0.5]$, is zero when $i\ne k$, $j \ne \ell$, (ii) ${\bm \Omega}_{ik}=0$ when $i\ne k$, $j = \ell$, and (iii) $[\bar{\bm S}^{-1}(f)]_{j \ell}=0$ , for every $f \in [0,0.5]$ when $i = k$, $j \ne \ell$. That is, we have a KPG ${\cal G} = {\cal G}_1 \otimes {\cal G}_2$ where the adjacency matrix of ${\cal G}_1$ is specified by the nonzero entries of $\bar{\bm S}^{-1}(f)$ $\forall f \in [0,0.5]$, and that of ${\cal G}_2$ follows from the nonzero entries of ${\bm \Omega}$.

Our objective is to learn the graph associated with $\{ {\bm Z}(t) \}$ under some sparsity constraints on ${\bm \Omega}$ and $\bar{\bm S}^{-1}(f), \, f \in [0,0.5]$. Since $\alpha \bar{\bm S}^{-1}(f) \otimes (\alpha^{-1} {\bm \Omega}) = \bar{\bm S}^{-1}(f) \otimes  {\bm \Omega}$, to resolve scaling ambiguity, we could normalize  $\|{\bm \Omega}\|_F=1$ or $\| \bar{\bm S}^{-1}(f_1) \; \cdots \; \bar{\bm S}^{-1}(f_M) ] \|_F =1$ for suitably placed $M$ frequencies in $(0,0.5)$; we will follow the latter as stated later in step 2 of Sec.\ \ref{FF}.
\vspace*{-0.17in}

\section{Penalized Negative Log-Likelihood} \label{pnll}
Given ${\bm z}(t)$ for $t=0, 1,2, \cdots , n-1$. Define the (normalized) DFT's ${\bm d}_z(f_m)$ and ${\bm D}_z(f_m)$ of ${\bm z}(t)$ and ${\bm Z}(t)$, respectively, as (recall $\iota = \sqrt{-1}$), 
\begin{align} 
   {\bm d}_z(f_m) = & \frac{1}{\sqrt{n}} \sum_{t=0}^{n-1} {\bm z}(t) \exp \left( - \iota 2 \pi f_m t \right) \, , \label{eqn050} \\
	{\bm D}_z(f_m) = & \frac{1}{\sqrt{n}} \sum_{t=0}^{n-1} {\bm Z}(t) \exp \left( - \iota 2 \pi f_m t \right) \, , \label{eqn053} \\
	   f_m = & m/n, \; m=0,1, \cdots , n-1. \label{eqn052}
\end{align}
Then ${\bm d}_z(f_m) = \mbox{vec} ({\bm D}_z(f_m))$.
It is established in \cite{Tugnait19d} (see also \cite{Tugnait22c}) that, for even $n$, the set of random vectors $\{{\bm d}_z(f_m)\}_{m=0}^{n/2}$ is a sufficient statistic for any inference problem based on dataset $\{ {\bm z}(t) \}_{t=0}^{n-1}$. Suppose  ${\bm S}_z(f_k)$ is locally smooth, so that ${\bm S}_z(f_k)$ is (approximately) constant over $K=2m_t+1$ consecutive frequency points $f_m$'s where $m_t$ is the half-window size; in our case, this assumption applies to $\bar{\bm S}(f_k)$. Pick $M =  \left\lfloor (\frac{n}{2}-m_t-1)/K \right\rfloor$ and
\begin{align} 
  \tilde{f}_k = & \big((k-1)K+m_t+1 \big) /n, \;\; \quad k \in [M], \label{window} 
\end{align}
yielding $M$ equally spaced frequencies $\tilde{f}_k$ in the interval $(0,0.5)$. We state the local smoothness assumption as assumption (A1).
\vspace*{0.04in}
\begin{itemize}
\setlength{\itemindent}{0.02in}
\item[(A1)] Assume that for $\ell = -m_t, -m_t+1, \cdots , m_t$,
\begin{align}  
   {\bm S}_z & (\tilde{f}_{k,\ell}) =  {\bm S}_z(\tilde{f}_k) \,  , \label{eqth1_160}  \\
	 \mbox{ where } \;  &
	  \tilde{f}_{k,\ell} = \big((k-1)K+m_t+1 + \ell \big)/n \, .
\end{align}
\end{itemize}
\vspace*{0.04in}

We will invoke \cite[Theorem 4.4.1]{Brillinger} for distribution of ${\bm d}_z(f_m)$. To this end we need assumption (A2).
\vspace*{0.04in}
\begin{itemize}
\setlength{\itemindent}{0.02in}
\item[(A2)] The matrix time series $\{ {\bm Z}(t) \}_{t=-\infty}^{\infty}$ is zero-mean stationary, Gaussian, satisfying $\sum_{\tau = -\infty}^\infty | [{\bm \Psi}( \tau )]_{k \ell} | < \infty$ for every $k, \ell \in [q]$.
\end{itemize}
\vspace*{0.04in}
\noindent By \cite[Theorem 4.4.1]{Brillinger}, under assumption (A1),  asymptotically (as $n \rightarrow \infty$), ${\bm d}_z(f_m)$, $m=1,2, \cdots , (n/2)-1$, ($n$ even), are independent proper, complex Gaussian ${\mathcal N}_c( {\bf 0}, {\bm S}_z(f_m))$ random vectors, respectively. Denote the joint probability density function of ${\bm d}_z(f_m)$, $m=1,2, \cdots , (n/2)-1$, as $f_{\bm{\mathcal D}}(\bm{\mathcal D})$ where ${\bm{\mathcal D}} = \{ {\bm D}_z(f_m) \}_{m=1}^{(n/2)-1}$. Then we have \cite{Tugnait19d, Tugnait22c}
\begin{align}  
   f_{\bm{\mathcal D}}(\bm{\mathcal D})  = & \prod_{k=1}^{M}  \left[ \prod_{\ell = -m_t}^{m_t}
	   \frac{\exp \left(- g_{kl} - g_{kl}^\ast\right) }
						 { \pi^{pq} \, | {\bm B}_k |^{1/2} 
						  | {\bm B}_k^\ast |^{1/2} }
		     \right]  \, , \label{eqn210} \\
  g_{kl} = & \frac{1}{2}{\bm d}_z^H(\tilde{f}_{k,\ell}) 
         \big( \bar{\bm S}^{-1}(\tilde{f}_k) \otimes {\bm \Sigma}^{-1} \big) 
			      {\bm d}_z(\tilde{f}_{k,\ell})  \, , \label{eqn212} \\
	{\bm B}_k = & \bar{\bm S}(\tilde{f}_k) \otimes {\bm \Sigma} \, . \label{eqn213}
\end{align}
Using $\mbox{tr} \big( {\bm A}^\top {\bm B} {\bm C} {\bm G}^\top \big) = (\mbox{vec}({\bm A}))^\top ({\bm G} \otimes {\bm B} ) \mbox{vec}({\bm C})$ and  parametrizing in terms of $\bm{\Phi}_k := \bar{\bm S}^{-1}(\tilde{f}_k)$ and ${\bm \Omega} = {\bm \Sigma}^{-1}$, we have 
\begin{align}  
 g_{kl} = & \frac{1}{2} {\bm D}_z^H(\tilde{f}_{k,\ell}) {\bm \Sigma}^{-1} {\bm D}_z(\tilde{f}_{k,\ell}) 
				 (\bar{\bm S}^{-1}(\tilde{f}_k))^\top \nonumber \\
			= & \frac{1}{2} {\bm D}_z^H(\tilde{f}_{k,\ell}) {\bm \Omega} {\bm D}_z(\tilde{f}_{k,\ell}) 
				 \bm{\Phi}_k^\ast \, . \label{eqn214}
\end{align}
Define $q \times (qM)$ matrix ${\bm \Gamma}$ as
\begin{align}  
  {\bm \Gamma} = & [ {\bm \Phi}_1 \;  {\bm \Phi}_2 \; \cdots \; {\bm \Phi}_M ] \, . \label{eqn400} 
\end{align}
Using  $|{\bm B}_k| = | \bar{\bm S}(\tilde{f}_k) \otimes {\bm \Sigma}| = 
| \bar{\bm S}(\tilde{f}_k) |^p  \, | {\bm \Sigma}|^q$,  up to some constants the negative log-likelihood follows from (\ref{eqn210}) as 
\begin{align} 
    - \frac{1}{KMpq} & \ln  f_{\bm{\mathcal D}}(\bm{\mathcal D}) 
	    \propto   G({\bm \Omega}, {\bm \Gamma}, {\bm \Gamma}^\ast ) \nonumber \\
						:= & - \frac{1}{p}  \ln (|{\bm \Omega}|) 
								- \frac{1}{2Mq} \sum_{k=1}^M \big( \ln (|\bm{\Phi}_k|) + \ln (|\bm{\Phi}_k^\ast| ) \big) \nonumber \\
			&		+ \frac{1}{2Mq} \sum_{k=1}^M \mbox{tr}( {\bm A}_k + {\bm A}_k^\ast )   \, , \label{eqn220} \\
  {\bm A}_k = & \frac{1}{Kp} \sum_{\ell = -m_t}^{m_t} 
        {\bm D}_z^H(\tilde{f}_{k,\ell}) {\bm \Omega} {\bm D}_z(\tilde{f}_{k,\ell}) 
				 \bm{\Phi}_k^\ast \, . \label{eqn225}              
\end{align}

In the high-dimension case, to enforce sparsity and to make the problem well-conditioned, we propose to minimize a penalized version ${\mathcal L} ({\bm \Omega}, {\bm \Gamma})$ w.r.t.\ ${\bm \Omega}$ and ${\bm \Gamma}$,
\begin{align}  
   {\mathcal L} ({\bm \Omega}, {\bm \Gamma})  = & G({\bm \Omega}, {\bm \Gamma}, {\bm \Gamma}^\ast) + P_p({\bm \Omega}) + P_q(\{ \bm{\Phi} \}) , \label{eqn230} \\
	   P_p({\bm \Omega})   = & \lambda_p \sum_{i \ne j}^p |{\bm \Omega}_{ij}|  = \lambda_p \| {\bm \Omega}^- \|_1 \, ,\\
	  P_q(\{ \bm{\Phi} \}) = &
	    \alpha \lambda_q \, \sum_{k=1}^M \; \sum_{i \ne j}^q 
		  \big| [ {\bm{\Phi}}_k ]_{ij} \big|  \nonumber \\
		& \;\;	+ (1-\alpha) \sqrt{M} \lambda_q \, \sum_{ i \ne j}^q \; \| {\bm{\Phi}}^{(ij)} \|  \, , \label{eqn231} \\
 {\bm{\Phi}}^{(ij)} := & [ [{\bm{\Phi}}_1 ]_{ij} \; [{\bm{\Phi}}_2 ]_{ij} \; \cdots \; [{\bm{\Phi}}_M ]_{ij}]^\top
     \in \mathbb{C}^M \, ,  \label{eqn232}
\end{align}
where $\{ \bm{\Phi} \} := \{ {\bm \Phi}_k \}_{k=1}^M$, $\alpha \in [0,1]$, $\lambda_p, \lambda_q > 0$ are tuning parameters, $P_p({\bm \Omega})$ is the lasso constraint, $P_q(\{ \bm{\Phi} \})$ is a sparse-group lasso sparsity constraint (cf.\ \cite{Friedman2010a, Friedman2010b, Yuan2006}) and $\sqrt{M}$ in $P_q(\{ \bm{\Phi} \})$ reflects number of group variables \cite{Yuan2006}. 
\vspace*{-0.12in}

\section{Optimization} \label{opt}
The objective function ${\mathcal L} ({\bm \Omega}, {\bm \Gamma})$ in (\ref{eqn230}) is biconvex: (strictly) convex in ${\bm \Gamma}$, ${\bm \Phi}_k \succ {\bm 0}$, for fixed ${\bm \Omega}$, and (strictly) convex in ${\bm \Omega}$, ${\bm \Omega} \succ {\bm 0}$, for fixed ${\bm \Gamma}$. As is a general approach for biconvex function optimization \cite{Gorski2007}, we will use an iterative and alternating minimization approach where we optimize w.r.t.\ ${\bm \Omega}$ with ${\bm \Gamma}$ fixed, and then optimize w.r.t.\ ${\bm \Gamma}$ with ${\bm \Omega}$ fixed at the last optimized value, and repeat the two optimizations (flip-flop). The algorithm is only guaranteed to converge to a local stationary point of ${\mathcal L} ({\bm \Omega}, {\bm \Gamma})$ \cite[Sec.\ 4.2.1]{Gorski2007}.  

With $\hat{{\bm \Gamma}} = [ \hat{\bm \Phi}_1 \;  \hat{\bm \Phi}_2 \; \cdots \; \hat{\bm \Phi}_M ]$ denoting the estimate of ${\bm \Gamma}$, fix ${\bm \Gamma} = \hat{\bm \Gamma}$ and let ${\mathcal L}_1({\bm \Omega})$ denote ${\mathcal L}({\bm \Omega}, \hat{\bm \Gamma})$ up to some irrelevant constants. We minimize ${\mathcal L}_1({\bm \Omega})$ w.r.t.\ ${\bm \Omega}$ to estimate $\hat{\bm{\Omega}}$, where
\begin{align}  
   & {\mathcal L}_1({\bm \Omega}) =  
	   - \frac{1}{p}  \ln (|{\bm \Omega}|) + 
		\frac{1}{p} \mbox{tr}\left( {\bm \Omega} \check{\bm \Theta} \right) 
		 + P_p({\bm \Omega}) \, , \label{eqn240}  \\
&	\check{\bm \Theta} =  \frac{1}{MKq} \sum_{k=1}^M \sum_{\ell = -m_t}^{m_t} \mbox{Re} \big\{
	{\bm D}_z(\tilde{f}_{k,\ell}) \hat{\bm \Phi}_k^\ast 
	 {\bm D}_z^H(\tilde{f}_{k,\ell}) \big \} \, . \label{eqn242}	    
\end{align}
Fix ${\bm \Omega}=\hat{\bm \Omega}$ and and let ${\mathcal L}_2({\bm \Gamma})$ denote ${\mathcal L}(\hat{\bm \Omega}, {\bm \Gamma})$ up to some irrelevant constants. We minimize ${\mathcal L}_2({\bm \Gamma})$ w.r.t.\ ${\bm \Gamma}$ to obtain estimate $\hat{\bm \Gamma}$, where
\vspace*{-0.1in}
\begin{align}  
    {\mathcal L}_2 &({\bm \Gamma}) =  
	   - \frac{1}{2Mq} \sum_{k=1}^M ( \ln (|\bm{\Phi}_k|) +\ln (|\bm{\Phi}_k^\ast|) )  \nonumber \\
	&	+ \frac{1}{2Mq} \sum_{k=1}^M \mbox{tr}\left( \tilde{\bm \Theta}_k \bm{\Phi}_k  +
						     \tilde{\bm \Theta}_k^\ast \bm{\Phi}_k^\ast \right)  
		+ P_q(\{ \bm{\Phi} \}) \, , \label{eqn250}  \\
&	\tilde{\bm \Theta}_k =  \frac{1}{Kp} \sum_{\ell = -m_t}^{m_t} 
        {\bm D}_z^\top(\tilde{f}_{k,\ell}) \hat{\bm \Omega} {\bm D}_z^\ast(\tilde{f}_{k,\ell}) \, .  \label{eqn252}	    
\end{align}
\vspace*{-0.04in}
Our optimization algorithm is as in Sec.\ \ref{FF}.
\vspace*{-0.16in}

\subsection{Flip-Flop Optimization} \label{FF}
\vspace*{-0.04in}
\begin{itemize}
\item[1.] Initialize $m=1$, ${\bm \Omega}^{(0)} = {\bm I}_p$, ${\bm \Phi}_k^{(0)} = {\bm I}_q$, $k \in [M]$.
\item[2.] Set $\hat{\bm \Omega} = {\bm \Omega}^{(m-1)}$ in (\ref{eqn252}). Use the iterative ADMM algorithm \cite{Boyd2010}, as outlined in \cite[Sec.\ 4]{Tugnait22c} and based on Wirtinger calculus \cite{Schreier10}, to minimize ${\mathcal L}_2 ({\bm \Gamma})$ (given by (\ref{eqn250})) w.r.t.\ ${\bm \Gamma}$ to obtain estimates ${\bm \Phi}_k^{(m)}$, $k \in [M]$, the $M$ component matrices of the estimate ${\bm \Gamma}^{(m)}$. Details are in Sec.\ \ref{ADMM1} and step II of Sec.\ \ref{PI}. Normalize ${\bm \Gamma}^{(m)} \leftarrow {\bm \Gamma}^{(m)}/ \|{\bm \Gamma}^{(m)}\|_F$ to resolve the scaling ambiguity. Let $m \leftarrow m+1$. 
\item[3.] Set $\hat{\bm \Gamma} = {\bm \Gamma}^{(m)}$ in  (\ref{eqn242}). Use the ADMM algorithm of \cite[Sec.\ III]{Tugnait21} (with $\alpha=1$ therein, no group-lasso penalty) to minimize ${\mathcal L}_1({\bm \Omega})$ w.r.t.\ ${\bm \Omega}$, to obtain estimate ${\bm \Omega}^{(m)}$. Details are in Sec.\ \ref{ADMM2} and step IV of Sec.\ \ref{PI}.
\item[4.] Repeat steps 2 and 3 until convergence. 
\end{itemize}
\vspace*{-0.16in}

\subsection{ADMM for Estimation of \texorpdfstring{${\bm \Gamma}$}{k} } \label{ADMM1}
\vspace*{-0.04in}
After variable splitting, the scaled augmented Lagrangian for minimization of ${\mathcal L}_2 ({\bm \Gamma})$ is \cite{Tugnait22c}
\begin{align*}
  {\mathcal L}_{2}^{aL} & ( \{ \bm{\Phi} \}, \{{\bm W} \}, \{{\bm U} \} ) 
	 = \frac{1}{2Mq} \sum_{k=1}^M \mbox{tr}\big( \tilde{\bm \Theta}_k \bm{\Phi}_k  +
						     \tilde{\bm \Theta}_k^\ast \bm{\Phi}_k^\ast \big) \\
	 & -\frac{1}{2Mq} \sum_{k=1}^M ( \ln (|\bm{\Phi}_k|) +\ln( |\bm{\Phi}_k^\ast| ) )+ {P}_q(\{ {\bm W} \})  \\
	& + \frac{\rho}{2} \sum_{k=1}^M \| \bm{\Phi}_k - {\bm W}_k + {\bm U}_k \|^2_F
\end{align*}
where $\{ {\bm U} \} = \{{\bm U}_k \}_{k=1}^M$ are dual variables, similarly $\{ {\bm W}_k \}_{k=1}^M$ are the ``split'' variables, $\rho >0$ is the penalty parameter, ${\bm U}_k, {\bm W}_k \in \mathbb{C}^{q \times q}$. 
Given the results $\{ \tilde{\bm{\Phi}}^{(i)} \}, \{{\bm W}^{(i)} \}, \{{\bm U}^{(i)} \}$ of the $i$th iteration, in the $(i+1)$st iteration, the ADMM algorithm executes the following three updates, given in Steps (a)-(c), until convergence. To distinguish between the estimates ${\bm \Gamma}^{(m)}$ and ${\bm \Phi}_k^{(m)}$ of the $m$th iteration of the flip-flip optimization and the estimate of the $i$th iteration of the ADMM algorithm, we use $\tilde{\bm{\Phi}}_k^{(i)}$ for the latter. \\
{\bf Step (a)}. $\{ \tilde{\bm{\Phi}}^{(i+1)} \} \leftarrow \arg \min _{\{ \bm{\Phi} \}} 
            {\mathcal L}_{2}^{aL} ( \{ \bm{\Phi} \}, \{{\bm W}^{(i)} \}, \{{\bm U}^{(i)} \} )$.  Up to some terms not dependent upon $\bm{\Phi}_k$'s \cite{Tugnait22c}
\begin{align*}
  {\mathcal L}_{2}^{aL} & ( \{ \bm{\Phi} \}, \{{\bm W}^{(i)} \}, \{{\bm U}^{(i)} ) \\
	 & = \frac{1}{2Mq} \sum_{k=1}^M {\mathcal L}_{2k}^{aL}  ( \bm{\Phi}_k , {\bm W}_k^{(i)} , {\bm U}_k^{(i)} ) \, , 
\end{align*}
\begin{align*}
	{\mathcal L}_{2k}^{aL} & ( \bm{\Phi}_k , {\bm W}_k^{(i)} , {\bm U}_k^{(i)} )
	  = -\ln (|\bm{\Phi}_k|) -\ln( |\bm{\Phi}_k^\ast| ) \\
		&   +\mbox{tr}\big( \tilde{\bm \Theta}_k \bm{\Phi}_k  + \tilde{\bm \Theta}_k^\ast \bm{\Phi}_k^\ast \big) 
		   + Mq \rho \| \bm{\Phi}_k - {\bm W}_k^{(i)} + {\bm U}_k^{(i)} \|^2_F \, ,
\end{align*}
that is, the objective function is separable in $k$. For each $k$, the solution is as follows \cite{Tugnait22c}.  Let ${\bm P}{\bm \Delta}{\bm P}^H$ denote the eigen-decomposition of the Hermitian $\tilde{\bm \Theta}_k - Mq \rho \left( {\bm W}_k^{(i)} - {\bm U}_k^{(i)} \right)$ with diagonal matrix ${\bm \Delta}$ consisting of the eigenvalues and ${\bm P}{\bm P}^H = {\bm P}^H{\bm P} = {\bm I}_q$. Then $\tilde{\bm{\Phi}}_k^{(i+1)} = {\bm P} \tilde{\bm \Delta} {\bm P}^H$ where $\tilde{\bm \Delta}$ is the diagonal matrix with $\ell$th diagonal element
\[
           [\tilde{\bm \Delta}]_{\ell \ell} 
 = \frac{ -[{\bm \Delta}]_{\ell \ell} + \sqrt{ ([{\bm \Delta}]_{\ell \ell})^2 + 4Mq \rho  } \, }{2Mq \rho } \, .
\] 
{\bf Step (b)}. Here we have 
\[
   \{ {\bf W}^{(i+1)} \} \leftarrow \arg \min _{\{ {\bm W} \}} 
            {\mathcal L}_{2}^{aL}  (\{ \tilde{\bm{\Phi}}^{(i+1)} \}, \{{\bm W} \}, \\ \{{\bm U}^{(i)} \} ).
\]
We update $\{ {\bm W}_k^{(i+1)} \}_{k=1}^M$ as the minimizer w.r.t.\ $\{ {\bm W} \}_{k=1}^M$ of 
\[
	\frac{\rho}{2} \sum_{k=1}^M \| {\bm W}_k - ( \tilde{\bm{\Phi}}_k^{(i+1)} + {\bm U}_k^{(i)} ) \|^2_F + {P}_q(\{ {\bm W} \}) .
\]
The solution follows from \cite[Lemma 1]{Tugnait22c}. Let ${\bm G}_k = \tilde{\bm{\Phi}}_k^{(i+1)} + {\bm U}_k^{(i)} \in \mathbb{C}^{q \times q}$ and let ${\bm G}^{(j \ell)} \in \mathbb{C}^M$ be defined as in (\ref{eqn232}), but based on ${\bm G}_k$'s. Then the update of $\{ {\bm W} \}$ is given by
\begin{align*}
	  [{\bm W}_k^{(i+1)}]_{j \ell} = & 
	    [{\bm G}_k]_{jj} \, , \quad\quad \mbox{if } j= \ell \\
	  [{\bm W}_k^{(i+1)}]_{j \ell} = &	\left( 1 - \frac{(1-\alpha)\lambda_q \sqrt{M}}
			{ \rho \| {\bm S}_F({\bm G}^{(j \ell)}, \, \alpha \lambda_q/\rho)\|}
			   \right)_+ \\
			& \quad \times		S_F \Big([{\bm G}^{(j \ell)}]_k, \frac{\alpha \lambda_q}{\rho} \Big) \, , \quad\quad \mbox{if } j \neq \ell \, ,					
\end{align*}
where $(b)_+ := \max(0,b)$, $S_F(b,\beta) := (1-\beta/|b|)_+ b$ (for complex scalar $b \neq 0$) is the soft-thresholding scalar operator, and $[{\bm S}_F({\bm a}, \beta)]_j = S(a_j,\beta)$ with $a_j = [{\bm a}]_j$, is the soft-thresholding vector operator. \\
{\bf Step (c)}. $\{ {\bm U}^{(i+1)} \} \leftarrow \{ {\bm U}^{(i)} \}  +
   \left( \{ \tilde{\bm{\Phi}}^{(i+1)} \}  - \{ {\bm W}^{(i+1)} \} \right)$.
\vspace*{-0.2in}

\subsection{ADMM for Estimation of \texorpdfstring{${\bm \Omega}$}{} } \label{ADMM2}
\vspace*{-0.04in}
Using variable splitting, consider 
\[
\min_{\bm{\Omega} \succ {\bm 0}, \bar{\bm W} }  \Big\{  \frac{1}{p} \big( {\rm tr} ( \check{\bm \Theta} \bm{\Omega}  ) 
          -   \ln(|\bm{\Omega}|) \big) + \lambda_p \| \bar{\bm W}^- \|_1   \Big\}
\]
subject to $\bm{\Omega} = \bar{\bm W}$. The scaled augmented Lagrangian for this problem is \cite{Boyd2010} 
\begin{align*}
					{\mathcal L}_{1}^{aL} & ( \bm{\Omega},  \bar{\bm W}, \bar{\bm U} ) = (1/p) \big( {\rm tr} ( \check{\bm \Theta} \bm{\Omega}  ) 
          -   \ln(|\bm{\Omega}|) \big)    \\
					& + \lambda_p \| \bar{\bm W}^- \|_1   
					 + \frac{\rho}{2}  \| \bm{\Omega} - \bar{\bm W} + \bar{\bm U}\|^2_F 
\end{align*}
where $\bar{\bm U}$ is the dual variable, and $\rho >0$ is the penalty parameter. Given the results $ \tilde{\bm{\Omega}}^{(i)}, \bar{\bm W}^{(i)}, \bar{\bm U}^{(i)}$ of the $i$th iteration, in the $(i+1)$st iteration, an ADMM algorithm executes the following three updates  until convergence: \\
{\bf Step (a)}. $\tilde{\bm{\Omega}}^{(i+1)} \leftarrow \arg \min_{\bm{\Omega}} \, 
  {\mathcal L}_{1}^{aL}( \bm{\Omega}, \bar{\bm W}^{(i)}, \bar{\bm U}^{(i)} )$. We choose $\bm{\Omega}$ to minimize
\[
{\rm tr} ( \check{\bm \Theta} \bm{\Omega}  ) 
          -   \ln(|\bm{\Omega}|) + \frac{p \rho}{2}  \| \bm{\Omega} - \bar{\bm W}^{(i)} + \bar{\bm U}^{(i)}\|^2_F.
\]
The solution is as follows \cite{Tugnait21}.  Let ${\bm Q}{\bm J}{\bm Q}^\top$ denote the eigen-decomposition of $\check{\bm \Theta} - p \rho \left( \bar{\bm W}^{(i)} - \bar{\bm U}^{(i)} \right)$ with diagonal matrix ${\bm J}$ consisting of the eigenvalues and ${\bm Q}{\bm Q}^\top = {\bm Q}^\top{\bm Q} = {\bm I}_q$. Then $\tilde{\bm{\Omega}}^{(i+1)} = {\bm Q} \tilde{\bm J} {\bm Q}^\top$ where $\tilde{\bm J}$ is the diagonal matrix with $\ell$th diagonal element
\[
           [\tilde{\bm J}]_{\ell \ell} 
 = \frac{ -[{\bm J}]_{\ell \ell} + \sqrt{ ([{\bm J}]_{\ell \ell})^2 + 4 p \rho  } \, }{2 p \rho } \, .
\] 
{\bf Step (b)}. $\bar{\bm W}^{(i+1)}  \leftarrow \arg \min _{ \bar{\bm W} } {\mathcal L}_{1}^{aL}( \tilde{\bm{\Omega}}^{(i+1)}, \bar{\bm W},  \bar{\bm U}^{(i)} )$. We update $\bar{\bm W}^{(i+1)}$ as the minimizer w.r.t.\ $\bar{\bm W}$ of 
\[
   \lambda_p \, \| \bar{\bm{W}}^- \|_1  
					+ \frac{\rho}{2}  \| \tilde{\bm{\Omega}}^{(i+1)} - \bar{\bm W} + \bar{\bm U}^{(i)} \|^2_F .
\]
The solution is soft thresholding given by \cite{Tugnait21} 
\begin{align*} 
    [\bar{\bm W}]_{jk}^{(i+1)} & = 
	 \left\{ \begin{array}{ll}
			    [ \tilde{\bm{\Omega}}^{(i+1)} - \bar{\bm U}^{(i)} ]_{jj} & \mbox{  if } j=k\\
					S_F([ \tilde{\bm{\Omega}}^{(i+1)} - \bar{\bm U}^{(i)} ]_{jk}, \frac{ \lambda_p}{\rho}) 
					   &  \mbox{ if } j \neq k \end{array} \right.
\end{align*}
where  $S_F()$ denotes soft-thresholding as in Sec.\ \ref{ADMM2}. \\
{\bf Step (c)}. $\bar{\bm U}^{(i+1)} \leftarrow \bar{\bm U}^{(i)}  +
   \left( \tilde{\bm{\Omega}}^{(i+1)} - \bar{\bm W}^{(i+1)} \right)$.
\vspace*{-0.2in}

\subsection{Practical Implementation} \label{PI}
Here we present our implementation of the algorithms of Secs.\ \ref{FF}-\ref{ADMM2} that was used in our numerical results. 
\begin{enumerate}[label = \Roman{*}.]
\item Parameters $\bar{\mu} =10$, $\tau_{rel} = \tau_{abs} = 10^{-4}$, $\tau_{ff} = 10^{-5}$, $m_{\max} = 20$, $i_{\max} = 100$ and $\rho^{(0)} =2$. Initialize $m=1$, ${\bm \Omega}^{(0)} = {\bm I}_p$, ${\bm \Phi}_k^{(0)} = {\bm I}_q$, $k \in [M]$.
\item For $m=1,2, \cdots , m_{\max}$, do steps III-IV.
\item Set $\hat{\bm \Omega} = {\bm \Omega}^{(m-1)}$ in (\ref{eqn252}). Pick $\rho = \rho^{(0)}$, $\tilde{\bm \Phi}_k^{(0)} = {\bm I}_q$ for $k \in [M]$. For $i=0,1, \cdots , i_{\max}$, do steps 1-6 below.
\begin{enumerate}[label = \arabic{*}.]
\item For $k \in [M]$, update ${\bm \Phi}_k$ as $\tilde{\bm \Phi}_k^{(i+1)}$ as in step (a), Sec.\ \ref{ADMM1}, then update ${\bm W}_k$ as ${\bm W}_k^{(i+1)}$ as in step (b), Sec.\ \ref{ADMM1}, and then update ${\bm U}_k$ as ${\bm U}_k^{(i+1)}$ as in step (c), Sec.\ \ref{ADMM1}, all with $\rho = \rho^{(i)}$.
\item Check for convergence following \cite[Sec.\ 4.1.5]{Tugnait22c}. Define the primal residual matrix ${\bm E}_{pri}^{(i+1)} \in \mathbb{C}^{q \times (qM)}$ at the $(i+1)$st iteration as
\begin{align*}
 {\bm E}_{pri}^{(i+1)} = & \Big[ \tilde{\bm \Phi}_1^{(i+1)} - {\bm W}_1^{(i+1)}, 
      \; \tilde{\bm \Phi}_2^{(i+1)} - {\bm W}_2^{(i+1)}, 
      \; \cdots , \\
			 & \quad\quad \tilde{\bm \Phi}_M^{(i+1)} - {\bm W}_M^{(i+1)}
           \Big]
\end{align*}
and the dual residual matrix ${\bm E}_{dual}^{(i+1)} \in \mathbb{C}^{q \times (qM)}$ at the $(i+1)$st iteration as
\begin{align*}
  {\bm E}_{dual}^{(i+1)} = & \rho^{(i)} \Big[  {\bm W}_1^{(i+1)} - {\bm W}_1^{(i)}, 
	  \; {\bm W}_2^{(i+1)} - {\bm W}_2^{(i)},\; \cdots , \\
       & \quad\quad   {\bm W}_M^{(i+1)} - {\bm W}_M^{(i)} \Big]  \, . 
\end{align*}
Let $e_1 = \|[\tilde{\bm \Phi}_1^{(i+1)} \; \tilde{\bm \Phi}_2^{(i+1)} \; \cdots \; \tilde{\bm \Phi}_M^{(i+1)}]\|_F$, 
$e_2 = \|[{\bm W}_1^{(i+1)} \; {\bm W}_2^{(i+1)} \; \cdots \; {\bm W}_M^{(i+1)}]\|_F$, 
$e_3 = \|[{\bm U}_1^{(i+1)} \; {\bm U}_2^{(i+1)} \; \cdots \; {\bm U}_M^{(i+1)}]\|_F$, 
$\tau_{pri} =  q \sqrt{M} \, \tau_{abs} + \tau_{rel} \, \max ( e_1, e_2 )$ and 
$\tau_{dual} =  q \sqrt{M}  \, \tau_{abs} + \tau_{rel} \,  e_3 / \rho^{(i)}$. 
If $\|{\bm E}_{pri}^{(i+1)}\|_F \le \tau_{pri}$ and $\|{\bm E}_{dual}^{(i+1)}\|_F \le \tau_{dual}$, the convergence criterion is met. If the convergence criterion is met or if $i+1 > i_{\max}$, exit to step IV after setting ${\bm \Phi}_k^{(m)} = \tilde{\bm \Phi}_k^{(i+1)}$, $k \in [M]$, and then normalizing ${\bm \Gamma}^{(m)} \leftarrow {\bm \Gamma}^{(m)}/ \|{\bm \Gamma}^{(m)}\|_F$, else continue,
\item Update variable penalty parameter $\rho$ as
\[
  \rho^{(i+1)} = \left\{ \begin{array}{ll} 2 \rho^{(i)} & 
	            \mbox{if  } \|{\bm E}_{pri}^{(i+1)}\|_F > \bar{\mu} \|{\bm E}_{dual}^{(i+1)} \|_F \\  
	           \rho^{(i)} /2 & \mbox{if  } \|{\bm E}_{dual}^{(i+1)} \|_F > \bar{\mu} \|{\bm E}_{pri}^{(i+1)}\|_F \\
						\rho^{(i)} & \mbox{otherwise}.  \end{array} \right.
\]
For $k \in [M]$, set ${\bm U}_k^{(i+1)} = {\bm U}_k^{(i)}/2$ if $\|{\bm E}_{pri}^{(i+1)}\|_F > \bar{\mu} \|{\bm E}_{dual}^{(i+1)} \|_F$ and ${\bm U}_k^{(i+1)} = 2 {\bm U}_k^{(i)}$ if $\|{\bm E}_{dual}^{(i+1)} \|_F > \bar{\mu} \|{\bm E}_{pri}^{(i+1)}\|_F$.
\item Set $i  \leftarrow i+1$ and return to step 2.
\end{enumerate} 
\item Set $\hat{\bm \Gamma} = {\bm \Gamma}^{(m)}$ in  (\ref{eqn242}). Pick $\rho = \rho^{(0)}$, $\tilde{\bm \Omega}^{(0)} = {\bm I}_p$. For $i=0,1, \cdots , i_{\max}$, do steps i-v below.
\begin{enumerate}[label = \roman{*}.]
\item Update ${\bm \Omega}$ as $\tilde{\bm \Omega}^{(i+1)}$ as in step (a), Sec.\ \ref{ADMM2}, then update $\bar{\bm W}$ as $\bar{\bm W}^{(i+1)}$ as in step (b), Sec.\ \ref{ADMM2}, and then update $\bar{\bm U}$ as $\bar{\bm U}^{(i+1)}$ as in step (c), Sec.\ \ref{ADMM2}, all with $\rho = \rho^{(i)}$.
\item Check for convergence following \cite[Sec.\ II-A]{Tugnait21}. Define the primal residual matrix ${\bm H}_{pri}^{(i+1)} =   \tilde{\bm \Omega}^{(i+1)} - \bar{\bm W}^{(i+1)}$ and the dual residual matrix ${\bm H}_{dual}^{(i+1)} =  \rho^{(i)} \big[  \bar{\bm W}^{(i+1)} - \bar{\bm W}^{(i)} \big]$ where ${\bm H}_{pri}^{(i+1)}, {\bm H}_{dual}^{(i+1)} \in \mathbb{C}^{p \times p}$. Let
\begin{align*}
	\tau_{pri} = & p\, \tau_{abs} + \tau_{rel} \, 
	        \max ( \|\tilde{\bm \Omega}^{(i+1)}\|_F , \|\tilde{\bm W}^{(i+1)}\|_F ) \\
  \tau_{dual} = & p  \, \tau_{abs} + \tau_{rel} \,  \| \tilde{\bm U}^{(i+1)} \|_F/ \rho^{(i)} \, .
\end{align*}
If $\|{\bm H}_{pri}^{(i+1)}\|_F \le \tau_{pri}$ and $\|{\bm H}_{dual}^{(i+1)}\|_F \le \tau_{dual}$, the convergence criterion is met. If the convergence criterion is met or if $i+1 > i_{\max}$, exit to step V after setting ${\bm \Omega}^{(m)} = \tilde{\bm \Omega}^{(i+1)}$, else continue.
\item Update variable penalty parameter $\rho$ as
\[
  \rho^{(i+1)} = \left\{ \begin{array}{ll} 2 \rho^{(i)} & 
	            \mbox{if  } \|{\bm H}_{pri}^{(i+1)}\|_F > \bar{\mu} \|{\bm H}_{dual}^{(i+1)} \|_F \\  
	           \rho^{(i)} /2 & \mbox{if  } \|{\bm H}_{dual}^{(i+1)} \|_F > \bar{\mu} \|{\bm H}_{pri}^{(i+1)}\|_F \\
						\rho^{(i)} & \mbox{otherwise}  \end{array} \right.
\]
Set $\bar{\bm U}^{(i+1)} = \bar{\bm U}^{(i)}/2$ if $\|{\bm H}_{pri}^{(i+1)}\|_F > \bar{\mu} \|{\bm H}_{dual}^{(i+1)} \|_F$ and $\bar{\bm U}^{(i+1)} = 2 \bar{\bm U}^{(i)}$ if $\|{\bm H}_{dual}^{(i+1)} \|_F > \bar{\mu} \|{\bm H}_{pri}^{(i+1)}\|_F$.
\item Set $i  \leftarrow i+1$ and return to step ii.
\end{enumerate} 
\item Check for convergence of the flip-flop algorithm. If $\|{\bm \Gamma}^{(m)} -  {\bm \Gamma}^{(m-1)}\|_F/\|{\bm \Gamma}^{(m-1)}\|_F \le \tau_{ff}$ and $\|{\bm \Omega}^{(m)} -  {\bm \Omega}^{(m-1)}\|_F/\|{\bm \Omega}^{(m-1)}\|_F \le \tau_{ff}$, or $m > m_{\max}$, go to step VI, else set $m \leftarrow m+1$ and return to step III.
\item The final estimates are given by $\hat{\bm \Omega} = {\bm \Omega}^{(m)}$ and $\hat{\bm \Gamma} = {\bm \Gamma}^{(m)}$, and
${\cal E}_{\hat{\bm \Omega}} = \{ (i,j) \, : \, |[\hat{\bm \Omega}]_{ij}| > 0 \}$ and ${\cal E}_{\hat{\bm \Gamma}} = \{ (i,j) \, : \, \|\hat{\bm \Phi}^{(ij)} \| > 0 \} $ are the estimated edgesets for ${\bm \Omega}$ and ${\bm \Gamma}$ respectively.
\end{enumerate}

{\it Remark 1}. We terminate the flip-flop optimization (step V) when relative improvements in new updates of both ${\bm \Omega}^{(m)}$ and ${\bm \Gamma}^{(m)}$ are below the threshold $\tau_{ff}$, or the maximum number of iterations in $m$ is reached. The ADMM algorithms are terminated when both primary and dual residuals are below the respective tolerances, or the maximum number of iterations in $i$ is reached; here we follow \cite[Sec.\ 3.3.1]{Boyd2010} (see also \cite{Tugnait21} and \cite{Tugnait22c}). The variable penalty $\rho^{(i)}$ follows the recommendations in \cite[Sec.\ 3.4.1]{Boyd2010}. 
The most expensive computation in Sec.\ \ref{ADMM1} is in step (a) requiring the eigen-decomposition of $M$ $q \times q$ matrices, with computational complexity $ {\cal O}(M q^3)$. Similarly, the most expensive computation in Sec.\ \ref{ADMM2} is in step (a) requiring the eigen-decomposition of a $p \times p$ matrix, with computational complexity $ {\cal O}(p^3)$. Thus the overall computational complexity of our proposed approach is ${\cal O}(M q^3 + p^3)$.
 $\;\; \Box$
\vspace*{-0.2in}

\subsection{BIC for selection of $\lambda_p$, $\lambda_q$ (and $\alpha$)}  \label{BIC}
\vspace*{-0.04in}
Given $n$, $K$ and $M$, the Bayesian information criterion (BIC) is given by (see also \cite{Tugnait22c})
\vspace*{-0.06in}
\begin{align}
 {\rm BIC}&(\lambda_p, \lambda_q, \alpha) =  -2KMq \ln (|\hat{\bm \Omega}|) \nonumber \\
  & \; + 2Kp  \sum_{k=1}^M \Big( -\ln (|\hat{\bm{\Phi}}_k|)  
	 + p^{-1} \, \mbox{Re}\big(\mbox{tr} ( \hat{\bm A}_k )  \big) \Big) \nonumber \\
	& \;	+ \ln (2 K M) \big( |\hat{\bm \Omega}|_0/2  + \sum_{k=1}^M |\hat{\bm{\Phi}}_k |_0 \big) \label{eqnBIC}
	\vspace*{-0.16in}
\end{align}
where $\hat{\bm A}_k$ is given by (\ref{eqn225}) with ${\bm \Omega}$ and $\bm{\Phi}_k$ therein replaced with $\hat{\bm \Omega}$ and $\hat{ \bm{\Phi}}_k$, respectively, $| {\bm H} |_0$ denotes number of nonzero elements in ${\bm H}$, $2KM$ is total number of real-valued measurements in frequency-domain and  $2K$ is the number of real-valued measurements per frequency point, with total $M$ frequencies in $(0, 0.5)$. A general expression for BIC is $-2\,$log-likelihood$+$(number of model parameters)$\times$log(number of data points). The expression in (\ref{eqnBIC}) follows by using $\{{\bm d}_z(f_m)\}_{m=1}^{(n/2)-1)}$ as complex-valued data in frequency-domain whose log-likelihood is given by (\ref{eqn220}). We count each complex value as two real values, both for data points and for parameters (entries of $\hat{\bm{\Phi}}_k$), and also use the fact that $\hat{\bm \Omega}$ is symmetric and $\hat{\bm{\Phi}}_k $ is Hermitian.

Pick $\alpha$, $\lambda_q$ and $\lambda_p$ to minimize BIC. In our simulations we fixed $\alpha = 0.05$ and then picked $\lambda_q$ and $\lambda_p$ over a grid of values, as follows. We search over $\lambda_q \in [\lambda_{q\ell} , \lambda_{qu}]$ and $\lambda_p \in [\lambda_{p\ell} , \lambda_{pu}]$ selected via a heuristic as in \cite{Tugnait21}. Find the smallest $\lambda_q$ and $\lambda_p$, labeled $\lambda_{qsm}$ and $\lambda_{psm}$, for which we get a no-edge model; then we set $\lambda_{qu}= \lambda_{qsm}/2$ and $\lambda_{q\ell} = \lambda_{qu}/10$; similarly for $\lambda_{pu}$ and $\lambda_{p \ell}$.  
\vspace*{-0.15in}

\section{Theoretical Analysis} \label{consist}
\vspace*{-0.06in}
Now we provide sufficient conditions for local convergence in the Frobenius norm of the Kronecker-decomposable inverse PSD estimators to the true value or a scaled version of it.  
First some notation. The true values of ${\bm \Omega}$, ${\bm \Sigma}$ and $\bar{\bm S}(f)$ will be denoted as ${\bm \Omega}^\diamond$, ${\bm \Sigma}^\diamond$ and $\bar{\bm S}^\diamond(f)$, respectively. Therefore, ${\bm \Omega}^\diamond = ({\bm \Sigma}^\diamond)^{-1}$. Since we use $\bm{\Phi}_k := \bar{\bm S}^{-1}(\tilde{f}_k)$, we have $\bm{\Phi}_k^\diamond := \bar{\bm S}^{-\diamond}(\tilde{f}_k)$ (where ${\bm A}^{-\diamond} = ({\bm A}^{\diamond})^{-1}$). Therefore, in this notation, ${\bm d}_z(f_m) \sim {\mathcal N}_c( {\bf 0}, {\bm S}_z^\diamond(f_m))$ and ${\bm S}_z^\diamond(f_m) =  \bar{\bm S}^\diamond(f_m) \otimes {\bm \Sigma}^\diamond$. Also in this section, we replace $\hat{\bm \Phi}_k$'s in (\ref{eqn242}) with ${\bm \Phi}_k$'s and still use the notation $\check{\bm \Theta}$ for the sum (\ref{eqn242}) and the notation ${\mathcal L}_1({\bm \Omega})$ for (\ref{eqn240}), and similarly, we replace $\hat{\bm \Omega}$ in (\ref{eqn252}) with ${\bm \Omega}$ and still use the notation $\tilde{\bm \Theta}$ for the sum (\ref{eqn252}) and ${\mathcal L}_2({\bm \Gamma})$ for (\ref{eqn250}).

We follow \cite{Lyu2020} in first considering the solution to the unpenalized population objective function (i.e., expectation of $G({\bm \Omega}, \{ \bm{\Phi} \}, \{ \bm{\Phi}^\ast \})$ given by \ref{eqn220}). We have
\begin{align}  
   & \bar{G}({\bm \Omega}, \{ \bm{\Phi} \}, \{ \bm{\Phi}^\ast \}) 
	 = E\{ G({\bm \Omega}, \{ \bm{\Phi} \}, \{ \bm{\Phi}^\ast \}) \} \nonumber \\
			& \quad =  - \frac{1}{p}  \ln (|{\bm \Omega}|) 
			- \frac{1}{2Mq} \sum_{k=1}^M \Big[ \ln (|\bm{\Phi}_k|)+ \ln (|\bm{\Phi}_k^\ast| ) \nonumber \\
	& \quad\quad - \frac{1}{p} \big( \mbox{tr}( \bar{\bm S}^\diamond_k {\bm \Phi}_k)
	  + \mbox{tr}( \bar{\bm S}^\diamond_k {\bm \Phi}_k)^\ast \big)
				 \mbox{tr}( {\bm \Sigma}^\diamond {\bm \Omega} ) \Big]
		 \, , \label{aeqn220} 
\end{align}
where we have used the facts $ \bar{\bm S}^\diamond_k = \bar{\bm S}^\diamond (\tilde{f}_{k})$, 
\begin{align}
\vspace*{-0.1in}
&	E\{ \mbox{tr}( {\bm A}_k )\} =  
 \frac{1}{Kp} \sum_{\ell = -m_t}^{m_t} 
        \mbox{tr}( E\{ {\bm d}_z(\tilde{f}_{k,\ell}) {\bm d}_z^H(\tilde{f}_{k,\ell}) \} 
				 ({\bm \Phi}_k \otimes {\bm \Omega}) ) \nonumber \\
& \quad = \frac{1}{p} 
        \mbox{tr}( ( \bar{\bm S}^\diamond_k \otimes {\bm \Sigma}^\diamond) 
				 ({\bm \Phi}_k \otimes {\bm \Omega}) )  = \frac{1}{p} 
        \mbox{tr}( ( \bar{\bm S}^\diamond_k {\bm \Phi}_k ) \otimes ({\bm \Sigma}^\diamond {\bm \Omega}) ) \nonumber \\
& \quad = p^{-1} \mbox{tr}( \bar{\bm S}^\diamond_k {\bm \Phi}_k) 
				 \mbox{tr}( {\bm \Sigma}^\diamond {\bm \Omega} )  \, . \label{aeqn222}	
\vspace*{-0.1in}
\end{align}
Define
\begin{align}
\vspace*{-0.06in}  
   \bar{\bm \Omega}({\bm \Gamma}) & = \arg \min_{\bm \Omega} \bar{G}({\bm \Omega}, \{ \bm{\Phi} \}, \{ \bm{\Phi}^\ast \}) \, ,
	            \label{aeqn250} \\
\bar{\bm \Gamma}({\bm \Omega}) & = \arg \min_{\bm \Gamma} \bar{G}({\bm \Omega}, \{ \bm{\Phi} \}, \{ \bm{\Phi}^\ast \})  
    \label{aeqn252} 
		\vspace*{-0.06in}
\end{align}
where $\bar{\bm \Gamma}({\bm \Omega}) =  [ \bar{\bm \Phi}_1({\bm \Omega}) \;  \bar{\bm \Phi}_2({\bm \Omega}) 
\; \cdots \; \bar{\bm \Phi}_M({\bm \Omega}) ]$. \\
{\it Theorem 1}. If  $\sum_{k=1}^M \big( \mbox{tr}( \bar{\bm S}^\diamond_k {\bm \Phi}_k)
	  + \mbox{tr}( \bar{\bm S}^\diamond_k {\bm \Phi}_k)^\ast \big) \, \neq \, 0$, then 
\begin{align}  
    \bar{\bm \Omega}({\bm \Gamma}) & = 
		  \frac{2Mq}{\sum_{k=1}^M \big( \mbox{tr}( \bar{\bm S}^\diamond_k {\bm \Phi}_k)
	  + \mbox{tr}( \bar{\bm S}^\diamond_k {\bm \Phi}_k)^\ast \big)} \, {\bm \Omega}^\diamond \, , \label{aeqn254}
\end{align}
and if $\mbox{tr}( {\bm \Sigma}^\diamond {\bm \Omega} ) \, \neq \, 0$, then for $k \in [M]$,
\begin{align}  
    \bar{\bm \Phi}_k({\bm \Omega}) & = 
		  \frac{p}{\mbox{tr}( {\bm \Sigma}^\diamond {\bm \Omega} )} \, (\bar{\bm S}^\diamond_k)^{-1} = 
		  \frac{p}{\mbox{tr}( {\bm \Sigma}^\diamond {\bm \Omega} )} \, \bm{\Phi}_k^\diamond  \quad \bullet \label{aeqn256}
\end{align}

Theorem 1 shows that the unpenalized population objective function yields true values up to a constant scalar. Notice that $\bar{\bm \Omega}({\bm \Gamma}) = {\bm \Omega}^\diamond$ if ${\bm \Gamma} = {\bm \Gamma}^\diamond$, and similarly, $\bar{\bm \Phi}_k({\bm \Omega}) = {\bm \Phi}_k^\diamond$, $k=1,2, \cdots, M$, if ${\bm \Omega} = {\bm \Omega}^\diamond$. 

We now turn to penalized data-based objective function ${\mathcal L} ({\bm \Omega}, {\bm \Gamma})$ which is minimized alternatingly as ${\mathcal L}_2 ({\bm \Gamma}) $ w.r.t.\ ${\bm \Phi}_k$'s and as ${\mathcal L}_1({\bm \Omega})$ w.r.t.\ ${\bm \Omega}$. Here in addition to assumptions (A1) and (A2), we assume
\begin{itemize}
\setlength{\itemindent}{0.02in}
\item[(A3)] Define the true edgesets ${\mathcal S}_q = \{ \{i,j\} ~:~ [(\bar{\bm S}^\diamond)^{-1}(f)]_{ij} \not\equiv 0, ~i\ne j,  ~ 0 \le f \le 0.5, ~ i,j \in [q] \}$ and ${\mathcal S}_p = \{ \{i,j\} ~:~ [{\bm \Omega}^\diamond]_{ij} \neq 0, ~i\ne j,  ~ i,j \in [p] \}$, where $\bar{\bm S}^\diamond(f) $ denotes DTFT of ${\bm \Psi}( \tau )$ and ${\bm \Omega}^\diamond = ({\bm \Sigma}^\diamond)^{-1}$ denotes the true value of ${\bm \Omega}$. Assume that number of nonzero elements in the true edgesets ${\mathcal S}_q$ and ${\mathcal S}_p$ are upperbounded as $|{\mathcal S}_q| \le s_{q}$ and $|{\mathcal S}_p| \le s_{p}$.

\item[(A4)] The minimum and maximum eigenvalues of $q \times q$ PSD $\bar{\bm S}^\diamond(f)  \succ {\bm 0}$  satisfy 
$0 < \beta_{q,\min}  \le \min_{f \in [0,0.5]} \phi_{\min}(\bar{\bm S}^\diamond(f))$ and $\max_{f \in [0,0.5]} \phi_{\max}(\bar{\bm S}^\diamond(f)) \le \beta_{q,\max} < \infty$. Similarly, $0 < \beta_{p,\min}  \le \phi_{\min}({\bm \Sigma}^\diamond) \le \phi_{\max}({\bm \Sigma}^\diamond)  \le \beta_{p,\max} < \infty$. Here $\beta_{\cdot,\min}$ and $\beta_{\cdot,\max}$ are not functions of $n, \, p \, , q$.
\end{itemize}

Theorem 2 establishes bounds on estimation errors of local minimizers $\hat{\bm \Omega}({\bm \Gamma})$ and $\hat{\bm \Gamma}({\bm \Omega})$ of ${\mathcal L}_1({\bm \Omega})$ and ${\mathcal L}_2 ({\bm \Gamma}) $, respectively. We now explicitly allow $p$, $q$, $M$, $K$, $s_p$, $s_q$, $\lambda_p$ and $\lambda_q$ to be functions of sample size $n$, denoted as $p_n$, $q_n$, $M_n$, $K_n$, $s_{pn}$, $s_{qn}$, $\lambda_{pn}$ and $\lambda_{qn}$, respectively. (In the appendices we do not do so to keep the notation simple.) First we define some variables. For $\tau > 2$, define
\begin{align} 
\gamma_p & = 0.1/\beta_{p,\max}\, , \label{eqn412c} \\ 
	C_{1q} & = \frac{2}{\sqrt{\ln(M_n^{1/\tau} q_n)}} + \sqrt{2 \tau + \frac{2 \, \ln(16)}{\ln(M_n^{1/\tau} q_n)}} \, , \label{eqn410b} 
\end{align}
\begin{align} 
C_{0q} & = 16 C_{1q} (1+\gamma_p \beta_{p,\max}) \beta_{q,\max}  \, , \label{eqn410} \\
	\gamma_q & = 0.1/\beta_{q,\max}\, , \label{eqn410c} \\
		C_{1p} & = \sqrt{\frac{2}{\ln(p_n)}} + \sqrt{\tau + \frac{\ln(4)}{\ln(p_n)}} \, , \label{eqn412b} \\
	 C_{0p} & = 8 C_{1p} \, (2+\gamma_q \beta_{q,\max}) \beta_{p,\max} \, , \label{eqn412} \\
	  r_{qn} & = \sqrt{M_n(q_n + s_{qn}) \, \ln (M_n^{1/\tau} q_n ) / ( K_n p_n ) } =o(1) \, , \label{eqn415}  \\
	  r_{pn} & = \sqrt{(p_n + s_{pn}) \, \ln ( p_n ) / (M_n K_n q_n )} =o(1) \, . \label{eqn417}   
\end{align}

{\it Theorem 2}. Let $\tau > 2$. \\
(i) Let ${\cal B}({\bm \Gamma}^\diamond) = \{ {\bm \Gamma}: \, \|{\bm \Gamma} - {\bm \Gamma}^\diamond \|_F \le \gamma_q, \; 
{\bm \Phi}_k ={\bm \Phi}_k^H \succ {\bm 0} \}$ and  
$\hat{\bm \Omega}({\bm \Gamma}) = \arg \min_{\{{\bm \Omega}: \, {\bm \Gamma} \in {\cal B}({\bm \Gamma}^\diamond) \}} 
         {\mathcal L}_1({\bm \Omega})$. Suppose $\lambda_{pn}$ satisfies 
\begin{align}  
    \frac{C_{0p}}{p_n}  \sqrt{\frac{\ln (p_n )}{M_n K_n q_n}} & \le \lambda_{pn}   
		\le \frac{C_{0p} }{p_n} 
		 \sqrt{ \left(1+ \frac{p_n}{s_{pn}} \right) \frac{\ln (p_n )}{M_n K_n q_n} }  \, . \label{eqn452}
\end{align}
Let $N_p := \arg \min_n \big\{ n : \, r_{pn} \le \beta_{p,\min}/(34 C_{0p}) \big\}$.
Then under assumptions (A1)-(A4), for $n > N_p$, $\hat{\bm \Omega}({\bm \Gamma})$ satisfies
\begin{align}  
  \| \hat{\bm \Omega}({\bm \Gamma}) - \bar{\bm \Omega}({\bm \Gamma}) \|_F & 
	   \le \frac{17 C_{0p}}{\beta_{p,\min}^2} \, r_{pn}  \label{eqn455}
\end{align}
with probability greater than $1 - \frac{1}{p^{\tau-2}} - 4 p^2 e^{ - KqM}$. \\
(ii) Let ${\cal B}({\bm \Omega}^\diamond) = \{ {\bm \Omega}: \, \|{\bm \Omega} - {\bm \Omega}^\diamond \|_F \le \gamma_p, \; 
{\bm \Omega} ={\bm \Omega}^\top \succ {\bm 0} \}$ and $\hat{\bm \Gamma}({\bm \Omega}) = \arg \min_{ \{{\bm \Gamma}: \, {\bm \Omega} \in {\cal B}({\bm \Omega}^\diamond) \}} {\mathcal L}_2({\bm \Gamma})$. Suppose $\lambda_{qn}$ satisfies 
\begin{align}  
   & \frac{C_{0q}}{M_n q_n} \, \sqrt{d_n} \le \lambda_{qn} \le \frac{C_{0q} }{M_n q_n } 
		 \sqrt{ \left(1+ \frac{q_n}{s_{qn}} \right) d_n }  \, , \label{eqn462}
\end{align}
where $d_n = \ln (M_n^{1/\tau}q_n ) / ( K_n p_n)$. 
Let $N_q := \arg \min_n \big\{ n : \, r_{qn} \le \beta_{q,\min}/(34 C_{0q}) \big\}$.
Then under assumptions (A1)-(A4), for $n > N_q$ and $\alpha \in [0,1]$, $\hat{\bm \Gamma}({\bm \Omega})$ satisfies
\begin{align} 
  \| \hat{\bm \Gamma}({\bm \Omega}) - \bar{\bm \Gamma}({\bm \Omega}) \|_F & 
	   \le \frac{17 C_{0q}}{\beta_{q,\min}^2} \, r_{qn}  \label{eqn465}
\end{align}
with probability greater than $1 - \frac{1}{q^{\tau -2}} - 16 M q^2 e^{- Kp/2} $ $\quad \bullet$

{\it Remark 2}. Theorem 2 helps determine how to choose $M_n$ and $K_n$ so that for given  $n$, $s_{pn}$, $s_{qn}$, $q_n$ and $p_n$, $\lim_{n \rightarrow \infty} r_{pn} =0$ and $\lim_{n \rightarrow \infty} r_{qn} =0$, and moreover, how fast can $s_{pn}$ and $s_{qn}$ grow with $n$ and still have $r_{qn}$ and $r_{pn}$ $\downarrow 0$. Since $K_n M_n \approx n/2$, if one picks $K_n = {\cal O}(n^\mu)$, then $M_n = {\cal O}(n^{1-\mu})$ for some $0 < \mu < 1$. We assume $p_n$ and $q_n$ are of the same order. (i) First consider the case ${\cal O}(p_n) = {\cal O}(p_n+s_{pn})= {\cal O}(q_n) = {\cal O}(q_n+s_{qn})$, which, for example, is true for chain graphs. Also, take ${\cal O}(p_n) \propto n^\nu $ for some $\nu > 0$. Then $r_{pn} = {\cal O}(\sqrt{\ln(n) / n})  \rightarrow 0$ as $n \rightarrow \infty$, and $r_{qn} = {\cal O}(\sqrt{\ln(n) / n^{2\mu-1}}) \rightarrow 0$ as $n \rightarrow \infty$ if $\mu > 0.5$. This holds for any $\nu > 0$. If $\mu = \frac{3}{4}$, then $r_{qn} = {\cal O}(\sqrt{\ln(n)}/n^{1/4}) > r_{pn}$. If $\mu = \frac{2}{3}$, then $r_{qn} = {\cal O}(\sqrt{\ln(n)}/n^{1/6}) > r_{pn}$. (ii) Now suppose ${\cal O}(p_n) = {\cal O}(q_n) \propto n^\nu$ for some $\nu > 0$, but ${\cal O}(s_{pn}) = {\cal O}(s_{qn}) \propto n^{2\nu} = {\cal O}(p_n^2)$, which is true for Erd\"{o}s-R\`{e}nyi graphs, e.g. Then $r_{pn} = {\cal O}(\sqrt{\ln(n) / n^{1-\nu}})  \rightarrow 0$ as $n \rightarrow \infty$ if $\nu < 1$, and $r_{qn} = {\cal O}(\sqrt{\ln(n) / n^{2\mu-1-\nu}}) \rightarrow 0$ as $n \rightarrow \infty$ if $2\mu - \nu > 1$. Clearly $\nu=1$ does not work. Suppose $\nu =0.25$ and $\mu = 0.75$. Then $r_{pn} = {\cal O}(\sqrt{\ln(n)}/n^{0.375})$ and $r_{qn} = {\cal O}(\sqrt{\ln(n)}/n^{1/8})$.
 $\;\; \Box$
	
{\it Remark 3}. The values of $\gamma_p$ and $\gamma_q$ specified in (\ref{eqn412c}) and (\ref{eqn410c}), respectively, are used in the proofs of Theorem 2(ii) (see after (\ref{apeqn620})) and Theorem 2(i) (see (\ref{apeqn515})), respectively. One can enlarge $\gamma_p$ and $\gamma_q$ to $\gamma_p = 0.1 \sqrt{p_n}/\beta_{p,\min} $ and $\gamma_q= 0.1 \sqrt{M_n q_n}/\beta_{q,\min}$, respectively, and the proofs and the other results remain unchanged and valid. Enlarging these values implies that the balls ${\cal B}({\bm \Gamma}^\diamond)$ and ${\cal B}({\bm \Omega}^\diamond)$ specified in Theorem 2 are larger, signifying larger convergence regions for initialization of ${\bm \Gamma}$ and ${\bm \Omega}$. However, this would slow the convergence rates from $\| \hat{\bm \Omega}({\bm \Gamma}) - \bar{\bm \Omega}({\bm \Gamma}) \|_F = {\cal O}_P(r_{pn})$ and $\| \hat{\bm \Gamma}({\bm \Omega}) - \bar{\bm \Gamma}({\bm \Omega}) \|_F = {\cal O}_P(r_{qn})$ to $\| \hat{\bm \Omega}({\bm \Gamma}) - \bar{\bm \Omega}({\bm \Gamma}) \|_F = {\cal O}_P(\sqrt{p_n} \, r_{pn})$ and$\| \hat{\bm \Gamma}({\bm \Omega}) - \bar{\bm \Gamma}({\bm \Omega}) \|_F = {\cal O}_P(\sqrt{M_n {q_n}} \, r_{qn})$, respectively.   $\;\; \Box$

{\it Theorem 3}. Assume $\| {\bm \Omega}^\diamond \|_F =1$. \\
(i) Define $\hat{\bm \Omega} = \hat{\bm \Omega}({\bm \Gamma})/ \|\hat{\bm \Omega}({\bm \Gamma})\|_F$. 
Let $N_{2p} := \arg \min_n \big\{ n : \, r_{pn} \le \beta_{p,\min}^2 \, \|\bar{\bm \Omega}({\bm \Gamma})\|_F/(34 C_{0p})  \,\big\}$ and $\gamma_r = (\beta_{q,\max}+\beta_{q,\min})/\beta_{q,\min}$.  Under the assumptions of Theorem 2(i), for $n > \max\{N_p,N_{2p}\}$, $\hat{\bm \Omega}$ satisfies
\begin{align}  
  \| \hat{\bm \Omega} - {\bm \Omega}^\diamond \|_F & \le 4 \gamma_r \| \hat{\bm \Omega}({\bm \Gamma}) - \bar{\bm \Omega}({\bm \Gamma}) \|_F 
	   \le \frac{68 \gamma_r C_{0p}}{\beta_{p,\min}^2} \, r_{pn}   \label{eqn467}
\end{align} 
with probability greater than $1 - \frac{1}{p^{\tau-2}} - 4 p^2 e^{ - KqM}$. \\
(ii) Let $C_{2p} = 68\gamma_r \sqrt{p}  \beta_{p,\max} C_{0p} / \beta_{p,\min}^2$, $U_{1p} = p/(2 C_{2p})$,  $U_{2p} = 0.1 \beta_{p,\min}^2/(68 \gamma_r C_{0p} \beta_{p,\max})$, $N_{3p} := \arg \min_n \big\{ n : \, r_{pn} \le \max\{U_{1p}, U_{2p}\} \big\}$ and $C_{2q} = 2 C_{2p} \| {\bm \Gamma}^\diamond \|_F  / p$.  Let $\hat{\bm \Gamma}(\hat{\bm \Omega}) = \arg \min_{ \{{\bm \Gamma}: \, {\bm \Omega} = \hat{\bm \Omega} \in {\cal B}({\bm \Omega}^\diamond) \}} {\mathcal L}_2({\bm \Gamma})$ where $\hat{\bm \Omega}$ is as in Theorem 3(i). 
Under the assumptions of Theorem 2, for $n > \max\{N_p, N_{2p}, N_q, N_{3p}\}$ and $\alpha \in [0,1]$, $\hat{\bm \Gamma}(\hat{\bm \Omega})$ satisfies
\begin{align} 
  \| \hat{\bm \Gamma}(\hat{\bm \Omega}) - {\bm \Gamma}^\diamond \|_F & 
	   \le \frac{17 C_{0q}}{\beta_{q,\min}^2} \, r_{qn}  + C_{2q} r_{pn} \label{eqn469}
\end{align}
with probability greater than $1 - \frac{1}{p^{\tau-2}} - 4 p^2 e^{ - KqM} - \frac{1}{q^{\tau -2}} - 16 M q^2 e^{- Kp/2} $ $\quad \bullet$
\vspace*{-0.1in}

\begin{table*}
\vspace*{-0.1in}
\caption{\small{\it $F_1$ scores, TPR, 1-TNR and timing per run for fixed tuning parameters, for the synthetic data example, averaged over 100 runs. The entries $**$ denote no simulations done for these parameters.}} \label{table1} 
\vspace*{-0.15in}
\begin{center}
\begin{tabular}{ccccccc}   \hline\hline
 $n $ & 64 & 128 &  256 & 512 & 1024 & 2048 \\  \hline\hline
\multicolumn{7}{c}{Proposed Approach:  $F_1$ scores $\pm \sigma$ when $\lambda$'s are selected to minimize BIC } \\ \hline
$M$=2  &  0.5163 $\pm$ 0.1530  &  0.5660 $\pm$ 0.1580  & 0.6440 $\pm$ 0.1709  &
            0.7061 $\pm$ 0.1147  &   0.7018 $\pm$ 0.1176  &   0.7190  $\pm$ 0.1217   \\  
$M$=3  &  0.5111 $\pm$ 0.1705  &  0.5876 $\pm$ 0.1560  & 0.6969 $\pm$ 0.1421  &
            0.7266 $\pm$ 0.1159  &   0.7322 $\pm$ 0.1031  &   0.7474  $\pm$ 0.1000   \\ 
$M$=4  &  0.5454 $\pm$ 0.1852  &  0.6470 $\pm$ 0.1489  & 0.7106 $\pm$ 0.1465  &
            0.7376 $\pm$ 0.1011  &   0.7446 $\pm$ 0.1069  &   0.7524  $\pm$ 0.1047   \\  
$M$=5  & 0.5977 $\pm$ 0.1717  &  0.6609 $\pm$ 0.1465  & 0.7049 $\pm$ 0.1367  &
            0.7253 $\pm$ 0.1104  &   0.7401 $\pm$ 0.1028  &   0.7467  $\pm$ 0.1002   \\ 
$M$=6  & $**$  &  0.6277 $\pm$ 0.1353  & 0.6773 $\pm$ 0.1379  &
            0.7115 $\pm$ 0.1172  &   0.7343 $\pm$ 0.1025  &   0.7369  $\pm$ 0.0971   \\
$M$=8  & $**$  & $**$   & $**$ &
            0.7016 $\pm$ 0.1071  &   0.7366 $\pm$ 0.1013  &   0.7365  $\pm$ 0.0974   \\
$M$=10  & $**$  &  $**$  & $**$  &
            0.7123 $\pm$ 0.1117  &   0.7319 $\pm$ 0.1044  &   0.7367  $\pm$ 0.1022   \\ \hline\hline
\multicolumn{7}{c}{Proposed Approach: $F_1$ scores $\pm \sigma$ when  $\lambda$'s are selected to maximize $F_1$ score } \\ \hline
$M$=2  &  0.6826 $\pm$ 0.1440  &  0.6954 $\pm$ 0.1588  & 0.7485 $\pm$ 0.1632  &
            0.8026 $\pm$ 0.1139  &   0.8032 $\pm$ 0.1588  &   0.8440  $\pm$ 0.1184   \\  
$M$=3  &  0.6984 $\pm$ 0.1383  &  0.7322 $\pm$ 0.1730  & 0.8055 $\pm$ 0.1383  &
            0.8293 $\pm$ 0.1190  &   0.8372 $\pm$ 0.1442  &   0.8670  $\pm$ 0.1295   \\ 
$M$=4  &  0.7041 $\pm$ 0.1355  &  0.7364 $\pm$ 0.1646  & 0.8074 $\pm$ 0.1434  &
            0.8282 $\pm$ 0.1169  &   0.8401 $\pm$ 0.1197  &   0.8633  $\pm$ 0.1397   \\  
$M$=5  &  0.6652 $\pm$ 0.1664  &  0.7309 $\pm$ 0.1431  & 0.8072 $\pm$ 0.1466  &
            0.8411 $\pm$ 0.1158  &   0.8451 $\pm$ 0.1251  &   0.8637  $\pm$ 0.1314 \\ 
$M$=6  & $**$  &  0.7218 $\pm$ 0.1490   & 0.8089 $\pm$ 0.1324  &
            0.8252 $\pm$ 0.1206  &   0.8433 $\pm$ 0.1282  &   0.8583  $\pm$ 0.1396   \\
$M$=8  & $**$  &  $**$  & $**$  &
            0.8329 $\pm$ 0.1130  &   0.8382 $\pm$ 0.1221  &   0.8601  $\pm$ 0.1404   \\
$M$=10  & $**$  & $**$   & $**$  &
            0.8187 $\pm$ 0.1216  &   0.8286 $\pm$ 0.1525  &   0.8496  $\pm$ 0.1406    \\ \hline\hline
\multicolumn{7}{c}{IID modeling \cite{Leng2012, Tsiligkaridis2013, Yin2012}: $\lambda$'s are selected to maximize $F_1$ score } \\ \hline
$F_1$ scores $\pm \sigma$   &  0.4329 $\pm$ 0.1244  &  0.4230 $\pm$ 0.1208  & 0.4368 $\pm$ 0.1228  &
            0.4746 $\pm$ 0.1367  &   0.4483 $\pm$ 0.1180  &   0.4709  $\pm$ 0.1104  \\ \hline
timing (s) per run $\pm \sigma$   &  0.0051 $\pm$ 0.0011  &  0.0073 $\pm$ 0.0014  & 0.0111 $\pm$ 0.0020  &
            0.0195 $\pm$ 0.0031  &   0.0342 $\pm$ 0.0035  &   0.0640  $\pm$ 0.0050  \\ \hline\hline
\multicolumn{7}{c}{Proposed approach under model mismatch -- non-Gaussian ${\bm e}$ in (\ref{eqn128}) : $F_1$ scores $\pm \sigma$ when $\lambda$'s are selected to minimize BIC } \\ \hline
Exponential ${\bm e}$, $M$=4  &  0.5518 $\pm$ 0.1853  &  0.6565 $\pm$ 0.1728  & 0.7098 $\pm$ 0.1349  &
            0.7355 $\pm$ 0.0976  &   0.7514 $\pm$ 0.1141  &   0.7555  $\pm$ 0.0888  \\ \hline
Uniform ${\bm e}$, $M$=4  &  0.5434 $\pm$ 0.1772  &  0.6510 $\pm$ 0.1693  & 0.7137 $\pm$ 0.1364  &
            0.7400 $\pm$ 0.1043  &   0.7494 $\pm$ 0.1146  &   0.7537  $\pm$ 0.0982  \\ \hline\hline
\multicolumn{7}{c}{Proposed Approach:  TPR $\pm \sigma$ when $\lambda$'s are selected to maximize $F_1$ score } \\ \hline
$M$=2  &  0.6312 $\pm$ 0.1675  &  0.6420 $\pm$ 0.1541  & 0.6937 $\pm$ 0.1852  &
            0.7533 $\pm$ 0.1332  &   0.8146 $\pm$ 0.1187  &   0.8249  $\pm$ 0.1199   \\  
$M$=4  &  0.6793 $\pm$ 0.1493  &  0.7120 $\pm$ 0.1477  & 0.7595 $\pm$ 0.1529  &
            0.7919 $\pm$ 0.1307  &   0.8142 $\pm$ 0.1229  &   0.8836  $\pm$ 0.1021   \\  
$M$=6  & $**$  &  0.6711 $\pm$ 0.1529  & 0.7459 $\pm$ 0.1608  &
            0.8024 $\pm$ 0.1287  &   0.8162 $\pm$ 0.1215  &   0.8275  $\pm$ 0.1290   \\
$M$=10  & $**$  &  $**$  & $**$  &
            0.7867 $\pm$ 0.1269  &   0.8278 $\pm$ 0.1199  &   0.8504  $\pm$ 0.1177  \\ \hline\hline
\multicolumn{7}{c}{Proposed Approach:  1-TNR $\pm \sigma$ when $\lambda$'s are selected to maximize $F_1$ score } \\ \hline
$M$=2  &  0.0032 $\pm$ 0.0092  &  0.0033 $\pm$ 0.0090  & 0.0022 $\pm$ 0.0074  &
            0.0018 $\pm$ 0.0061  &   0.0049 $\pm$ 0.0157  &   0.0025  $\pm$ 0.0096   \\  
$M$=4  &  0.0041 $\pm$ 0.0118  &  0.0044 $\pm$ 0.0127  & 0.0020 $\pm$ 0.0074  &
            0.0021 $\pm$ 0.0097  &   0.0023 $\pm$ 0.0086  &   0.0043  $\pm$ 0.0174  \\  
$M$=6  & $**$  &  0.0035 $\pm$ 0.0116  & 0.0013 $\pm$ 0.0050  &
            0.0026 $\pm$ 0.0113  &   0.0027 $\pm$ 0.0120 &   0.0030  $\pm$ 0.0161   \\
$M$=10  & $**$  &  $**$  & $**$  &
            0.0025 $\pm$ 0.0113  &   0.0046 $\pm$ 0.0173  &   0.0040  $\pm$ 0.0173   \\ \hline\hline
\multicolumn{7}{c}{Proposed Approach:  timing (s) per run $\pm \sigma$ when $\lambda$'s are selected to minimize BIC } \\ \hline
$M$=2  &  0.1687 $\pm$ 0.0400  &  0.1688 $\pm$ 0.0418  & 0.1791 $\pm$ 0.1005  &
            0.1777 $\pm$ 0.0289  &   0.2166 $\pm$ 0.0322  &   0.3131  $\pm$ 0.1051   \\  
$M$=4  &  0.2294 $\pm$ 0.1650  &  0.2470 $\pm$ 0.1026  & 0.2284 $\pm$ 0.0846  &
            0.2278 $\pm$ 0.0392  &   0.2890 $\pm$ 0.1338  &   0.3627  $\pm$ 0.0494   \\  
$M$=6  & $**$  &  0.2738 $\pm$ 0.0903  & 0.2426 $\pm$ 0.0633  &
            0.2507 $\pm$ 0.0537  &   0.2944 $\pm$ 0.0902  &   0.3842  $\pm$ 0.0471   \\
$M$=10  & $**$  &  $**$  & $**$  &
            0.3040 $\pm$ 0.1400 &   0.3230 $\pm$ 0.0733  &   0.4285  $\pm$ 0.0890 \\ \hline\hline
\end{tabular} 
\end{center}
\vspace*{-0.2in}
\end{table*}

\section{Numerical Results} \label{NE}
We now present numerical results for both synthetic and real data to illustrate the proposed approach. In synthetic data examples the ground truth is known and this allows for assessment of the efficacy of various approaches. In real data examples where the ground truth is unknown, our goal is visualization and exploration of the linear conditional dependency structures underlying the data.
\vspace*{-0.1in}

\subsection{Synthetic Data} \label{NEsyn}
We use model (\ref{eqn128})-(\ref{eqn129}) to generate synthetic data where ${\bm \Psi}(\tau)$ is controlled via a vector autoregressive (VAR) model impulse response and ${\bm \Sigma}$ is determined via an Erd\"{o}s-R\`{e}nyi graph. We take $p=q=15$. Consider the impulse response ${\bm H}^{(r)}_i \in \mathbb{R}^{5 \times 5}$ generated as $ {\bm H}^{(r)}_i = \sum_{k=1}^3 {\bm A}^{(r)}_k {\bm H}^{(r)}_{i-k} + {\bm I}_5 \delta_i$, where ${\bm H}^{(r)}_i = 0$ for $i<0$, $\delta_i$ is the Kronecker delta, $r=1,2,3$, and  only 5\% of entries of ${\bm A}^{(r)}_i$'s are nonzero and the nonzero elements are independently and uniformly distributed over $[-0.8,0.8]$. We then check if the VAR(3) model is stable with all eigenvalues of the companion matrix $\le 0.95$ in magnitude; if not, we re-draw randomly till this condition is fulfilled. The impulse response ${\bm B}_i \in \mathbb{R}^{15 \times 15}$ in (\ref{eqn128}) is given by ${\bm B}_i = \mbox{block-diag} \{ {\bm H}^{(1)}_i \, ,  {\bm H}^{(2)}_i \, ,  {\bm H}^{(3)}_i \}$, for $0 \le i \le L=40$, otherwise it is set to zero. Thus ${\bm B}_i$'s in (\ref{eqn128}) have a block-diagonal structure with 3 blocks, each block is $5 \times 5$. In the Erd\"{o}s-R\`{e}nyi graph with $p=15$ nodes, the nodes are connected with probability $p_{er} =0.05$. In the upper triangular $\bar{\bm \Omega}$, $\bar{\bm \Omega}_{ij} =0$ if $\{i,j\} \not\in {\mathcal S}_p$, $\bar{\bm \Omega}_{ij}$ is uniformly distributed over $[-0.4,-0.1] \cup [0.1,0.4]$ if $\{i,j\} \in {\mathcal S}_p$, and $\bar{\bm \Omega}_{ii} =0.5$. With $\bar{\bm \Omega} = \bar{\bm \Omega}^\top$,  add $\kappa {\bm I}_p$ to  $\bar{\bm \Omega}$ with $\kappa$ picked to make minimum eigenvalue of $\bm{\Omega}=\bar{\bm \Omega} + \kappa {\bm I}_p$ equal to 0.5. Let ${\bm \Omega} = \tilde{\bm F} \tilde{\bm F}$ (matrix square-root), then ${\bm F} = \tilde{\bm F}^{-1}$ in (\ref{eqn128}). 
\vspace*{-0.1in}
\begin{figure}[ht]
  \centering
  \includegraphics[width=0.65\linewidth]{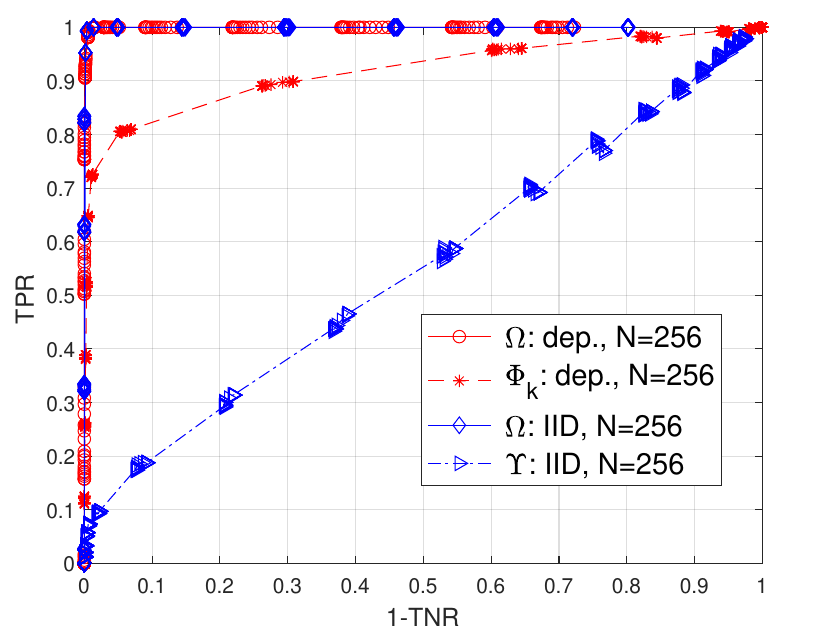} 
	\vspace*{-0.1in}
  \caption{\small{ROC curves: plots labeled ``IID'' are from the approach of \cite{Leng2012, Tsiligkaridis2013, Yin2012}, and the plots labeled ``dep.'' are from our proposed approach. TPR=true positive rate, TNR=true negative rate}}
  \label{fig1}
\end{figure}
\vspace*{-0.1in}

We applied our proposed approach with $n=256$, $M=2$, $K=63$ and compared with the approach of \cite{Leng2012} (which is also the approach of \cite{Tsiligkaridis2013, Yin2012}, all of whom assume i.i.d.\ observations and have two lasso penalties one each on ${\bm \Omega}$ and ${\bm \Upsilon}$, counterpart to our ${\bm \Phi}_k$). In our approach, we fix $\alpha = 0.05$ for all simulations and real data results. For fixed values of $\lambda_q$ and $\lambda_p$, using our proposed approach of Sec.\ \ref{PI}, we calculated the true positive rate (TPR) and false positive rate 1-TNR (where TNR is the true negative rate) over 100 runs, separately for ${\bm \Omega}$ and $\{{\bm \Phi}_k\}$/${\bm \Upsilon}$, based on the estimated edges. As we vary $\lambda_q$ and $\lambda_p$ over a wide range of values, we can compute the corresponding  pairs of estimated (1-TNR, TPR). The receiver operating characteristic (ROC) is shown in Fig.\ \ref{fig1} based on 100 runs, using the estimated (1-TNR, TPR). We repeat this method for the i.i.d.\ modeling approach of \cite{Leng2012, Tsiligkaridis2013, Yin2012}. Fig.\ \ref{fig1} shows that the i.i.d.\ modeling of \cite{Leng2012, Tsiligkaridis2013, Yin2012} is unable to capture the ``dependent'' edges (cf.\ (\ref{eqn125})) via ${\bm \Upsilon}$ whereas it has no issues with  ${\bm \Omega}$. Our approach works well for both components of the Kronecker product graph.

\begin{figure*}
\begin{subfigure}[b]{0.33\textwidth}
\begin{center}
\includegraphics[width=0.8\linewidth]{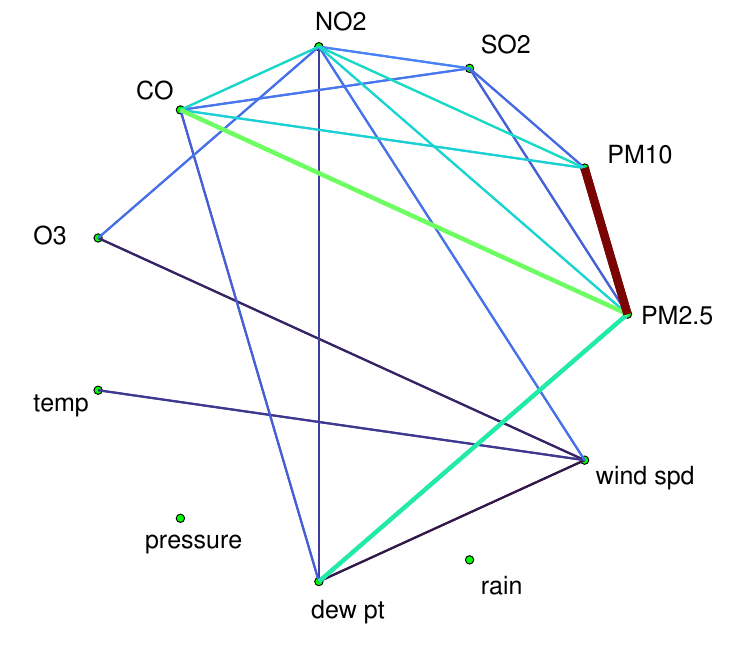}
\caption{Pollution graph: \\ proposed approach, M=4}
\end{center}
\end{subfigure}%
\begin{subfigure}[b]{0.33\textwidth}
\begin{center}
\includegraphics[width=0.8\linewidth]{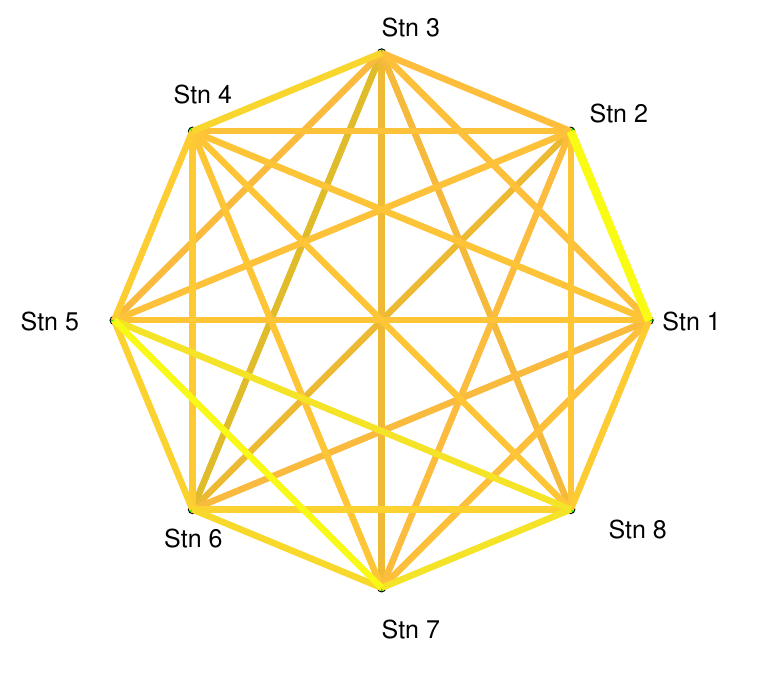}
\caption{Monitoring site graph: \\ proposed approach, M=4}
\end{center}
\end{subfigure}
\begin{subfigure}[b]{0.33\textwidth}
\begin{center}
\includegraphics[width=0.85\linewidth]{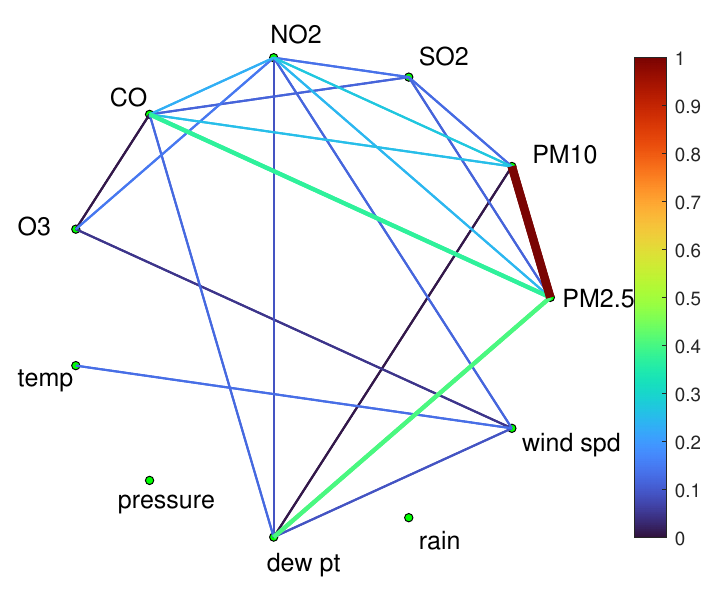}
\caption{Pollution graph: \\ proposed approach, M=3}
\end{center}
\end{subfigure}%
\newline %
\begin{subfigure}[b]{0.33\textwidth}
\begin{center}
\includegraphics[width=0.8\linewidth]{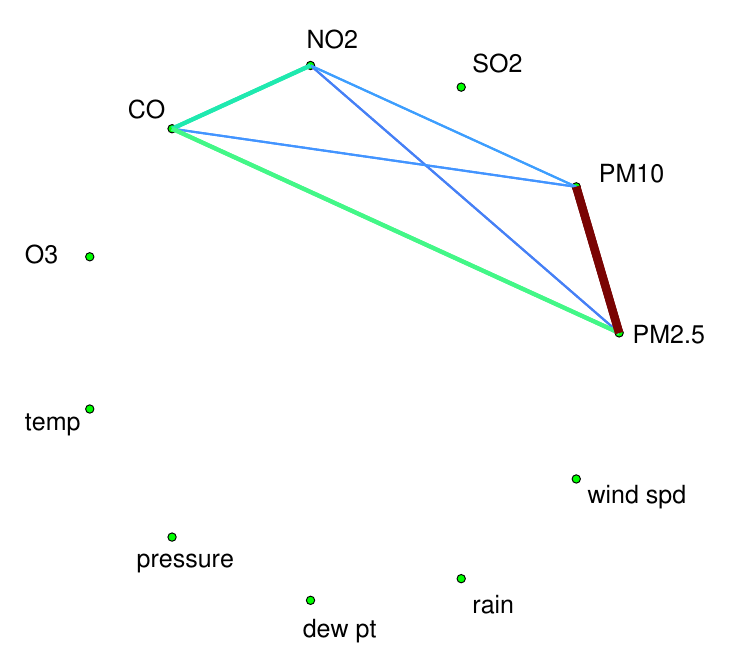}
\caption{Pollution graph: \\ proposed approach, M=5}
\end{center}
\end{subfigure}%
\begin{subfigure}[b]{0.33\textwidth}
\begin{center}
\includegraphics[width=0.8\linewidth]{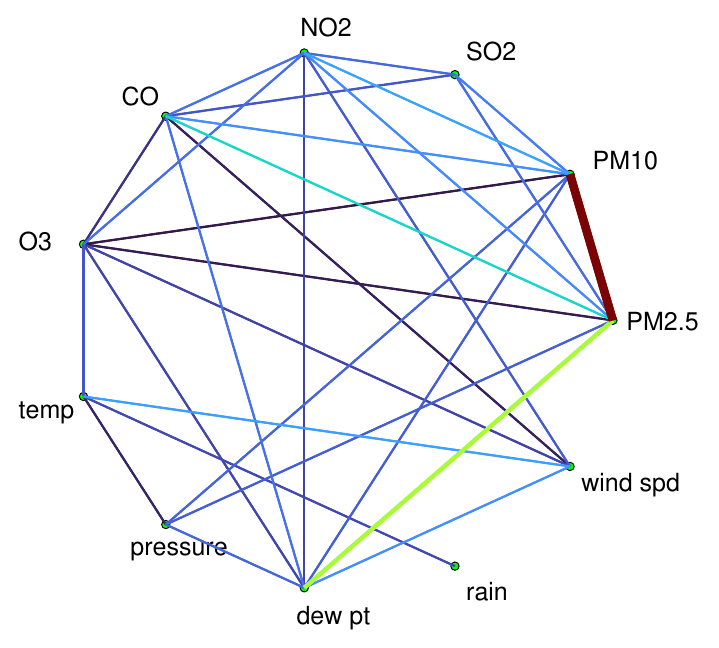}
\caption{Pollution graph: \\ i.i.d.\ modeling approach \cite{Leng2012, Tsiligkaridis2013, Yin2012}  } 
\end{center}
\end{subfigure}%
\begin{subfigure}[b]{0.33\textwidth}
\vspace*{-0.1in}
\begin{center}
\vspace*{-0.1in}
\includegraphics[width=0.85\linewidth]{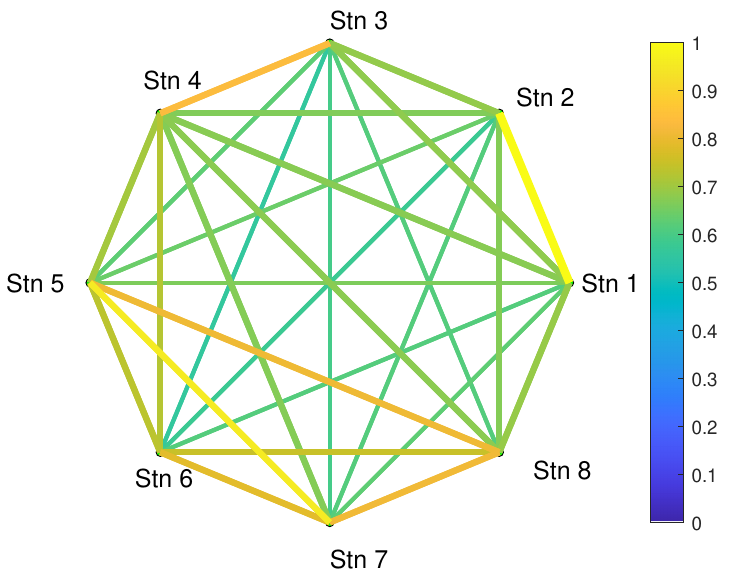}
\caption{Monitoring site graph: \\ i.i.d.\ modeling approach \cite{Leng2012, Tsiligkaridis2013, Yin2012}}
\end{center}
\end{subfigure}%
\vspace*{-0.05in}
\caption{\small{Pollution and site graphs for the Beijing air-quality dataset \cite{Zhang2017} for year 2013-14: 8 monitoring sites and 11 features ($p=8$, $q=11$, $n=364$). Number of distinct edges $=18, \, 28, \, 20, \, 6, \, 30, \, 28$ in graphs (a), (b), (c), (d), (e) and (f), respectively. Monitoring sites labeled Stn.\ 1-4 are the rural/suburban sites and those labeled Stn.\ 5-8 are the urban sites (see the text). For the pollution graph, estimated $\hat{\bm \Phi}^{(ij)}$ is the edge weight (normalized to have $\max_{i \ne j}\hat{\bm \Phi}^{(ij)}=1$) and for the site graph, estimated $|\hat{\Omega}_{ij}|$ is the edge weight (normalized to have $\max_{i \ne j}|\hat{\Omega}_{ij}|=1$). The edge weights are color coded (all pollution graphs share the same color legend, and similarly for the site graphs), in addition to the edges with higher weights being drawn thicker.}} \label{figreal}
\vspace*{-0.18in}
\end{figure*}

In Table \ref{table1} we show the results based on 100 runs under different parameter settings and samples sizes. Here we show the $F_1$score, TPR, 1-TNR and timing values for the overall graph (not the two Kronecker product components separately) along with the $\pm \sigma$ errors. All algorithms were run on a Window 10 Pro operating system with processor Intel(R) Core(TM) i7-10700 CPU @2.90 GHz with 32 GB RAM, using MATLAB R2023a. We take $n=64, 128, 256, 512, 1024, 2048$, and for our proposed approach, the corresponding  $m_t$ values leading to different $M$ values are $m_t= 7, 15, 31, 63, 127, 255 $ ($M=2$), $m_t= 4, 9, 20, 41, 84, 169 $ ($M=3$), $m_t= 3, 7, 14, 31, 63, 127 $ ($M=4$), $m_t= 2, 5, 12, 24, 50, 101 $ ($M=5$), $m_t= **, 4, 10, 20, 42, 84 $ ($M=6$), $m_t= **, **, ** , 15, 31, 63 $ ($M=8$), and $m_t= **, **, **, 12, 25, 50 $ ($M=10$). Here $**$ denotes that no simulation were performed for the corresponding sample size $n$ (since $K=2m_t+1$ is too small). We show the resulting $F_1$ scores under two different scenarios: when we use the proposed BIC parameter selection method (Sec.\ \ref{BIC}) and when $F_1$ score was selected based on $\lambda$ values that maximize the $F_1$ score. While the latter approach is not practical, it is presented to illustrate what is possible using the proposed approach and what may be ''lost'' when there are errors in the BIC parameter selection method. The number of unknown parameters being estimated are ${\cal O}(p^2 + M q^2)$, with ${\cal O}(p^2)$ for ${\bm \Omega}$ and ${\cal O}(M q^2)$ for $M$ ${\bm \Phi}_k$'s. We see that for a fixed $n$, at first the performance improves with increasing $M$, then it slowly declines as more parameters need to be estimated with increasing $M$. Increasing $M$ also reduces $K=2m_t+1$ since $KM \approx \frac{n}{2}$, which reduces the number of frequency-domain samples ($K$) for averaging for the $k$th model for ${\bm \Phi}_k$, $k \in [M]$ (see assumption (A1) in Sec.\ \ref{pnll}). Note also that by (\ref{eqn465}) of Theorem 2(ii), the error in estimating ${\bm \Phi}_k$'s $\propto r_{qn} \propto \sqrt{(Mq)/(Kp)}$. For a fixed $M$, the performance improves, in general, with increasing $n$ but more slowly for higher $n$'s. Higher $n$ values implies higher resolution in the frequency-domain and for fixed $M$, higher $n$ implies higher $K$ (and $m_t$), in which case assumption (A1) in Sec.\ \ref{pnll} may not hold. The TPR, 1-TNR and timing values are shown for selected $M$'s for the proposed approach where timing per run is for the $\lambda$ values picked by the BIC criterion. It is seen that increasing $M$ and/or $n$ leads to only a small increase in timing. 

In Table \ref{table1} we also show the performance of i.i.d.\ modeling approach of \cite{Leng2012, Tsiligkaridis2013, Yin2012},
 in terms of the $F_1$ score and timing. The i.i.d.\ modeling approach is significantly faster but the accuracy in edge detection in terms of the $F_1$ score is much poorer. Finally, to assess sensitivity to modeling errors such as violation of the Gaussianity assumption, we used either exponential or uniform ${\bm e}(t)$ in (\ref{eqn128}), both with zero-mean unit variance, instead of the assumed Gaussian ${\bm e}(t)$ in our model. The results are shown for $M=4$ and we see that the performance is robust w.r.t.\ violation of this assumption.
\vspace*{-0.1in}

\subsection{Real Data: Beijing air-quality dataset \cite{Zhang2017}} \label{NEreal}
Here we consider Beijing air-quality dataset \cite{Zhang2017, Chen2015}, downloaded from \url{https://archive.ics.uci.edu/dataset/501/beijing+multi+site+air+quality+data}. This data set includes hourly air pollutants data from 12 nationally-controlled air-quality monitoring sites in the Beijing area. The time period is from March 1st, 2013 to February 28th, 2017. The six air pollutants are PM$_{2.5}$, PM$_{10}$, SO$_2$, NO$_2$, CO, and O$_3$, and the meteorological data is comprised of five features: temperature, atmospheric pressure, dew point, wind speed, and rain; we did not use wind direction. Thus we have eleven ($=q$) features (pollutants and weather variables). We used data from 8 ($=p$) sites: 4 rural/suburban sites Changping, Dingling, Huairou, Shunyi, and 4 urban sites Aotizhongxin, Dongsi, Guanyuan, Gucheng (labeled Stn 1 through 8 in Fig.\ \ref{figreal}). The data are averaged over 24 hour period to yield daily averages. We used one year 2013-14 of daily data resulting in $n = 365$ days. Arranging stations as rows and features as columns, we have ${\bm Z}(t) \in \mathbb{R}^{8 \times 11}$, $t=1,2, \cdots , 365$. 
We pre-processed the data as follows. Given $j$th feature data ${\bm Z}_{ij}(t)$ at $i$th station, we transform it to $\bar{\bm Z}_{ij}(t) = \ln({\bm Z}_{ij}(t)/{\bm Z}_{ij}(t-1))$ for each $i$ and $j$, and then detrend it (i.e., remove the best straight-line fit). Finally, we scale the detrended scalar sequence to have a mean-square value of one. 
All temperatures were converted from Celsius to Kelvin to avoid negative numbers. If a value of a feature is zero (e.g., wind speed), we added a small positive number to it so that the log transformation is well-defined.  
\vspace*{-0.05in}

We applied our proposed approach with $M=4$, $K=45$ and $n=364$ ($p=8$, $q=11$) and compared it with the i.i.d.\ modeling approach of \cite{Leng2012, Tsiligkaridis2013, Yin2012}. The objective here is to compare the two approaches in estimation of the pollution (feature) graph and the site graph. The spatio-temporal data has a matrix structure and one is interested in learning two aspects of conditional dependencies: the relationship among the features via the pollution graph and the relationship among the sites via the site graph. We have not yet tested if our model assumptions apply to this dataset (this needs further theoretical analysis to devise suitable statistical tests, particularly in a high-dimensional setting), but it still seems to be useful to compare the results of our proposed approach and that of  \cite{Leng2012, Tsiligkaridis2013, Yin2012}. 
Fig.\ \ref{figreal}(a) shows the resulting graph for the air quality and environmental variables where $\{ i, j\} \in {\mathcal S}_q$ iff $\hat{\bm \Phi}^{(ij)} = (\sum_{k=1}^M | [\hat{\bm{\Phi}}_k]_{ij} |^2)^{1/2} > 0$ for $i \ne j$, and Fig.\ \ref{figreal}(b) shows the resulting graph for the sites around the Beijing area where $\{ i, j\} \in {\mathcal S}_p$ iff $|\hat{\Omega}_{ij}| > 0$ for $i \ne j$. Since all the sites are physically close to one another, it is not surprising that the site graph in Fig.\ \ref{figreal}(b) is fully connected. But we do see that the rural/suburban sites stn.\ 1 through stn.\ 4 have higher weight edges among the group and the urban sites stn.\ 5 through stn.\ 8 have higher weight edges among the urban group, with inter-group edge weights being slightly weaker (but fully connected). Automobile exhaust is the main cause of NO$_2$ which is likely to undergo a chemical reaction with Ozone O$_3$, thereby, lowering its concentration \cite{Chen2015}. This fact is captured by the edge between NO$_2$ and Ozone O$_3$ in Fig.\ \ref{figreal}(a). Cold, dry air from the north of Beijing reduces both dew point and PM$_{2.5}$ particle concentration in suburban areas while southerly wind brings warmer and more humid air from the more polluted south that elevates both dew point and PM$_{2.5}$ concentration \cite{Zhang2017}. This fact is captured by the edge between dew point and PM$_{2.5}$ in Fig.\ \ref{figreal}(a). The counterparts to Figs.\ \ref{figreal}(a) and \ref{figreal}(b) when using the i.i.d.\ modeling approach of \cite{Leng2012, Tsiligkaridis2013, Yin2012}, are shown in Figs.\ \ref{figreal}(e) and \ref{figreal}(f), respectively. While the site graph in Fig.\ \ref{figreal}(f) is fully connected and quite similar to the proposed approach's site graph in Fig.\ \ref{figreal}(b), the pollution graph in Fig.\ \ref{figreal}(e) far denser than the proposed approach's pollution graph in Fig.\ \ref{figreal}(a).
\vspace*{-0.03in}

We do not have any systematic approach for selection of $M$ for a given sample size $n$. Since $KM \approx \frac{n}{2}$, fixing $M$ fixes $K=2m_t+1$, and vice-versa. Using BIC to pick $M$ does not work as BIC always picks the smallest $M$. The synthetic data results presented in Table \ref{table1} show that the performance is not unduly sensitive  to the choice of $M$. To illustrate the sensitivity of the proposed approach in Beijing data case, we show the pollution graphs in Figs.\ \ref{figreal}(c) and \ref{figreal}(d) for the choice $M=3$ (K=59) and $M=5$ ($K=35$), respectively. There is not much difference between pollution graphs for $M=4$ and $M=3$, but that for $M=5$ is much sparser. This is consistent with the results of  Sec.\ \ref{NEsyn} on synthetic data. 
\vspace*{-0.08in}

\section{Conclusions} Sparse-group lasso penalized log-likelihood approach in frequency-domain with a Kronecker-decomposable PSD was investigated for matrix CIG learning for dependent time series. An ADMM-based flip-flop approach for iterative optimization of the bi-convex problem was presented. We provided sufficient conditions for consistency of a local estimator of inverse PSD. We illustrated our approach using numerical examples utilizing both synthetic and real data.  Lasso and related approaches are known to yield biased estimates \cite{Fan2009}. To remedy this, various non-convex penalties have been suggested \cite{Fan2009} and typically, lasso-based approaches provide the initial guess for iterative optimization. In the context of this paper, adaptive lasso has been used in \cite{Tugnait21} (the basis of the ADMM method of Sec.\ \ref{ADMM2}), and a log-sum penalty has been used in \cite{Tugnait2022} (which modifies \cite{Tugnait22c}, the basis for Sec.\ \ref{ADMM1}). Investigation of such non-convex penalties is left for future research. 
\vspace*{-0.1in}

\appendices
\section{Proof of Theorem 1} \label{append1}
With fixed ${\bm \Gamma}$, let $\bar{G}_1({\bm \Omega})$ denote $\bar{G}({\bm \Omega}, \{ \bm{\Phi} \}, \{ \bm{\Phi}^\ast \})$ up to some irrelevant constants. Then
\begin{align}  
    \bar{G}_1({\bm \Omega}) 
	 = & - \frac{1}{p}  \ln (|{\bm \Omega}|) + B \, \mbox{tr}( {\bm \Sigma}^\diamond {\bm \Omega} )
		 \, , \label{apeqn220} 
\end{align}
where $B = \mbox{tr}( \bar{\bm S}^\diamond_k {\bm \Phi}_k)^\ast / (2Mqp)$.
We have
\begin{align}  
   {\bm 0} = & \frac{\partial  \bar{G}_1({\bm \Omega}) }{\partial {\bm \Omega} }
	 =  - \frac{1}{p}  {\bm \Omega}^{-1} + B \,  {\bm \Sigma}^\diamond 
		 \, , \label{apeqn224} 
\end{align}
establishing (\ref{aeqn254}) if $B \neq 0$. The solution is unique since the Hessian of $\bar{G}_1({\bm \Omega})$, given by $\frac{1}{p} {\bm \Omega}^{-1} \otimes {\bm \Omega}^{-1}$, is positive definite at ${\bm \Omega} = \bar{\bm \Omega}({\bm \Gamma})$. Similarly, with fixed ${\bm \Omega}$, let $\bar{G}_2({\bm \Gamma})$ denote $\bar{G}({\bm \Omega}, \{ \bm{\Phi} \}, \{ \bm{\Phi}^\ast \})$ up to some irrelevant constants. Then
\begin{align}  
    \bar{G}_2({\bm \Gamma}) 
	 = & \sum_{k=1}^M \bar{G}_{2k}({\bm \Phi}_k)
		 \, , \label{apeqn226} 
\end{align}
\begin{align}  
   \bar{G}_{2k}({\bm \Phi}_k) & =   -  \ln (|\bm{\Phi}_k|) - \ln (|\bm{\Phi}_k^\ast|)  \nonumber \\
	 & \quad + \frac{1}{p} \big( \mbox{tr}( \bar{\bm S}^\diamond_k {\bm \Phi}_k)  
	   + \mbox{tr}( \bar{\bm S}^\diamond_k {\bm \Phi}_k)^\ast \big)
				 \mbox{tr}( {\bm \Sigma}^\diamond {\bm \Omega} ) \, . \label{apeqn228} 
\end{align}
The cost $\bar{G}_2({\bm \Gamma})$ is separable in $k$, ${\bm \Phi}_k$. We have
\begin{align}  
   {\bm 0} = & \frac{\partial  \bar{G}_{2k}({\bm \Phi}_k) }{\partial \bm{\Phi}_k^\ast }
	 =  - {\bm \Phi}_k^{-1} + \frac{1}{p}  \bar{\bm S}^\diamond_k \mbox{tr}( {\bm \Sigma}^\diamond {\bm \Omega} ) 
		 \, , \label{apeqn230} 
\end{align}
establishing (\ref{aeqn256}) if $\mbox{tr}( {\bm \Sigma}^\diamond {\bm \Omega} ) \neq 0$. Similar to \cite[Lemma 4]{Tugnait22c}, the Hessian of $\bar{G}_{2k}({\bm \Phi}_k)$ is positive definite at ${\bm \Phi}_k = \bar{\bm \Phi}_k({\bm \Omega})$. Therefore, the solution is unique. $\quad \Box$
\vspace*{-0.1in}

\section{Technical Lemmas and Proof of Theorem 2} \label{append2}
Lemma 1 is a restatement of \cite[Lemma S.1, Supplementary]{Lyu2020}. \\
{\it Lemma 1}. Assume that i.i.d.\ data ${\bm X}_i \in {\mathbb R}^{p \times q}$, $i=1,2, \cdots , n$, follows the matrix-valued normal distribution $\mathcal{MVN}({\bm 0}, {\bm \Sigma}^\diamond, {\bm \Psi}^\diamond)$, with ${\bm \Sigma}^\diamond \in {\mathbb R}^{p \times p}$, ${\bm \Psi}^\diamond \in {\mathbb R}^{q \times q}$, ${\bm \Sigma}^\diamond \succ {\bm 0}$ and ${\bm \Psi}^\diamond \succ {\bm 0}$, i.e., $\mbox{vec}({\bm X}_i) \, \sim \, {\cal N}_r \big( {\bm 0}, {\bm \Psi}^\diamond \otimes {\bm \Sigma}^\diamond \big)$. Assume that $\phi_{max}({\bm \Sigma}^\diamond) \le C_{1h} < \infty$ and $\phi_{max}({\bm \Psi}^\diamond) \le C_{2h} < \infty$ for some positive constants $C_{1h}$ and $C_{2h}$. For any symmetric positive-definite ${\bm \Omega} \in \mathbb{R}^{p \times p}$ such that $\| {\bm \Omega} - {\bm \Omega}^\diamond\|_F \le \gamma$, ${\bm \Omega}^\diamond =({\bm \Sigma}^\diamond)^{-1}$,  we have
\begin{align}
  P & \Big( \max_{i,j} \Big| \big[ \frac{1}{np} \sum_{i=1}^n {\bm X}_i^\top {\bm \Omega} {\bm X}_i -
	      \frac{1}{p} E\{ {\bm X}_i^\top {\bm \Omega} {\bm X}_i\} \big]_{ij} \Big| \ge \delta \Big) \nonumber \\
		& \le 4 q^2 \Big[ \exp\big\{ - \frac{np}{2} \big[ \frac{\delta}{8(1+\gamma C_{1h})C_{2h}} - \frac{2}{\sqrt{np}} 
		   \big]^2 \big\}  \nonumber \\
		& \quad\quad + \exp\big\{-\frac{np}{2} \big\} \Big]  \label{apeqn300}
\end{align}
for any $\delta > 16(1+\gamma C_{1h})C_{2h}/\sqrt{np} \;$ $\quad \bullet$ \\
The lower bound on $\delta$ follows from \cite[Lemma S.12, Supplementary]{Lyu2020} and (\ref{apeqn300}) is \cite[Eqn.\ (S.26), Supplementary]{Lyu2020} in our notation. 

Lemma 2 collects some useful results from \cite[Theorem 2.3.5]{Gupta1999}. \\
{\it Lemma 2}. Suppose ${\bm X} \sim \mathcal{MVN}({\bm 0}, {\bm \Sigma}, {\bm \Psi})$ where ${\bm X} \in \mathbb{R}^{p \times q}$, ${\bm \Sigma} \in {\mathbb R}^{p \times p}$, ${\bm \Psi} \in {\mathbb R}^{q \times q}$, i.e., $\mbox{vec}({\bm X}) \, \sim \, {\cal N}_r \big( {\bm 0}, {\bm \Psi} \otimes {\bm \Sigma} \big)$. Then
\begin{itemize}
\item[(i)] ${\bm X}^\top \sim \mathcal{MVN}({\bm 0}, {\bm \Psi}, {\bm \Sigma})$, i.e., $\mbox{vec}({\bm X}^\top) \, \sim \, {\cal N}_r \big( {\bm 0}, {\bm \Sigma} \otimes {\bm \Psi} \big)$.
\item[(ii)] For any ${\bm A} \in \mathbb{R}^{q \times q}$, $E\{ {\bm X} {\bm A} {\bm X}^\top \} = \mbox{tr}({\bm A}^\top {\bm \Psi}) \, {\bm \Sigma}$.
\item[(iii)] For any ${\bm B} \in \mathbb{R}^{p \times p}$, $E\{ {\bm X}^\top {\bm B} {\bm X} \} = \mbox{tr}({\bm B}^\top {\bm \Sigma}) \, {\bm \Psi}$.
\item[(iv)] For any ${\bm C} \in \mathbb{R}^{q \times p}$, $E\{ {\bm X} {\bm C} {\bm X} \} = {\bm \Sigma} {\bm C}^\top {\bm \Psi}$ $\quad \bullet$ 
\end{itemize}

{\it Lemma 3}. Suppose ${\bm X} \in \mathbb{C}^{p \times q}$, $\mbox{vec}({\bm X}) \, \sim \, {\cal N}_c \big( {\bm 0}, {\bm S} \otimes {\bm \Sigma} \big)$ where ${\bm \Sigma} \in {\mathbb R}^{p \times p}$, ${\bm S} \in {\mathbb C}^{q \times q}$, ${\bm \Sigma} = {\bm \Sigma}^\top \succ {\bm 0}$, ${\bm S} = {\bm S}^H \succ {\bm 0}$ and ${\bm S} = {\bm S}_r + j {\bm S}_i$ with ${\bm S}_r, {\bm S}_i \in \mathbb{R}^{q \times q}$. 
\begin{itemize}
\item[(i)] Let ${\bm X} = {\bm X}_r + j {\bm X}_i$, ${\bm X}_r, {\bm X}_i \in \mathbb{R}^{p \times q}$. Then 
\begin{equation}
  \tilde{\bm X} = [{\bm X}_r \; {\bm X}_i]  \sim \mathcal{MVN}({\bm 0}, {\bm \Sigma}, \tilde{\bm S})
\end{equation}
i.e., $\mbox{vec}(\tilde{\bm X}) \, \sim \, {\cal N}_r \big( {\bm 0}, \tilde{\bm S} \otimes {\bm \Sigma} \big)$, where
\begin{equation}
  \tilde{\bm S} = \frac{1}{2} \begin{bmatrix}
	    {\bm S}_r & -{\bm S}_i \\ {\bm S}_i & {\bm S}_r \end{bmatrix} \in \mathbb{R}^{2q \times 2q} \; .
\end{equation}
\item[(ii)] For any ${\bm \Omega} \in \mathbb{R}^{p \times p}$, $E\{ \tilde{\bm X}^\top {\bm \Omega} \tilde{\bm X} \} = \mbox{tr}({\bm \Omega}^\top {\bm \Sigma}) \, \tilde{\bm S}$.
\item[(iii)] For any ${\bm \Phi} \in \mathbb{C}^{q \times q}$, ${\bm \Phi} = {\bm \Phi}_r + j {\bm \Phi}_i = {\bm \Phi}^H$, ${\bm \Phi}_r, {\bm \Phi}_i \in \mathbb{R}^{q \times q}$,
\begin{equation}
E\{ \mbox{Re} \big( {\bm X} {\bm \Phi}^\ast {\bm X}^H \big) \} = 
  E\{ \tilde{\bm X} \tilde{\bm \Phi} \tilde{\bm X}^\top \big) \} = \mbox{tr}(\tilde{\bm \Phi}^\top \tilde{\bm S}) \, {\bm \Sigma}
\end{equation}
where
\begin{equation}
  \tilde{\bm \Phi} = \begin{bmatrix}
	    {\bm \Phi}_r & -{\bm \Phi}_i \\ {\bm \Phi}_i & {\bm \Phi}_r \end{bmatrix} \in \mathbb{R}^{2q \times 2q} \; \quad \bullet 
\end{equation} 
\end{itemize}
{\it Proof}. \begin{itemize}
\item[(i)] If ${\bm x} = \mbox{vec}({\bm X}) \, \sim \, {\cal N}_c \big( {\bm 0}, {\bm S} \otimes {\bm \Sigma} \big)$, then by \cite[Sec.\ 2.3]{Schreier10}, 
\begin{align}
  \tilde{\bm x} = & \mbox{vec}(\tilde{\bm X}) \, \sim \, {\cal N}_r \big( {\bm 0}, {\bm R} \big)
\end{align}
where, with ${\bm x} = {\bm x}_r + j {\bm x}_i$, ${\bm x}_r, {\bm x}_i \in \mathbb{R}^{pq}$,
\begin{align}
  {\bm R} = & = \begin{bmatrix} E\{ {\bm x}_r {\bm x}_r^\top \} &  E\{ {\bm x}_i {\bm x}_r^\top \} \\ 
	  E\{ {\bm x}_r {\bm x}_i^\top \} &  E\{ {\bm x}_i {\bm x}_i^\top \} \end{bmatrix}
	= \begin{bmatrix} {\bm R}_{rr} & {\bm R}_{ir} \\ {\bm R}_{ri} & {\bm R}_{ii} \end{bmatrix} \\
	{\bm R}_{rr} = & {\bm R}_{ii} \, , \quad {\bm R}_{ri} = - {\bm R}_{ri}^\top = {\bm R}_{ir}^\top \, .
\end{align} 
Now ${\bm R}_{rr} = \frac{1}{2} {\bm S}_r \otimes {\bm \Sigma} = {\bm R}_{ii}$ and ${\bm R}_{ri} = - \frac{1}{2} {\bm S}_i^\top \otimes {\bm \Sigma}$. Therefore, ${\bm R} = \tilde{\bm S} \otimes {\bm \Sigma}$, yielding the desired result.
\item[(ii)] It follows from Lemma 2(iii) and Lemma 3(i).
\item[(iii)] Since ${\bm \Phi} = {\bm \Phi}^H$, it follows that ${\bm \Phi}_r = {\bm \Phi}_r^\top$ and ${\bm \Phi}_i = -{\bm \Phi}_i^\top$. We have $\mbox{Re} \big( {\bm X} {\bm \Phi}^\ast {\bm X}^H \big) = \tilde{\bm X} \tilde{\bm \Phi} \tilde{\bm X}^\top $. Then the given expression for $E\{ \tilde{\bm X} \tilde{\bm \Phi} \tilde{\bm X}^\top \big) \}$ follows from Lemma 2(ii). $\quad \Box$
\end{itemize}

We now consider a tail bound on $\tilde{\bm \Theta}_k $ defined in (\ref{eqn252}). First we need Lemma 4. \\
{\it Lemma 4}. Given ${\bm S} \in {\mathbb C}^{q \times q}$ and $\tilde{\bm S}\in \mathbb{R}^{2q \times 2q}$ as in Lemma 3. Then $\tilde{\bm S} \succ {\bm 0}$ and $\phi_{max}(\tilde{\bm S}) = \frac{1}{2} \phi_{max}({\bm S})$. \\
{\it Proof}. If $\lambda$ is a an eigenvalue of ${\bm S}$, then for some ${\bm v} = {\bm v}_r + j {\bm v}_i \in \mathbb{C}^q$, ${\bm v}_r, {\bm v}_i \in \mathbb{R}^q$, we have ${\bm S} {\bm v} = \lambda {\bm v}$, where $\lambda$ is real positive since ${\bm S}$ is Hermitian, positive-definite. It then follows that 
\begin{align}
  \tilde{\bm S} & \begin{bmatrix} {\bm v}_r \\ {\bm v}_i \end{bmatrix}  
	   = \frac{1}{2} \lambda \begin{bmatrix} {\bm v}_r \\ {\bm v}_i \end{bmatrix} \, , \quad 
	\tilde{\bm S}  \begin{bmatrix} -{\bm v}_i \\ {\bm v}_r \end{bmatrix}  
	  = \frac{1}{2} \lambda \begin{bmatrix} -{\bm v}_i \\ {\bm v}_r \end{bmatrix}  \, . \label{apeqn340}
\end{align}
That is, each eigenvalue of ${\bm S}$ is also an eigenvalue of $2  \tilde{\bm S}$ with multiplicity two. This proves the desired result. 
$\quad \Box$ \\
{\it Lemma 5}. Under assumptions (A1) and (A2), for any symmetric positive-definite $\hat{\bm \Omega} \in \mathbb{R}^{p \times p}$ such that $\| \hat{\bm \Omega} - {\bm \Omega}^\diamond\|_F \le \gamma_p$, ${\bm \Omega}^\diamond =({\bm \Sigma}^\diamond)^{-1}$,  and $\tau > 2$, we have
\begin{align}
  P & \Big( \max_{k,i,j} \big| \big[ \tilde{\bm \Theta}_k^\ast -
	       E\{ \tilde{\bm \Theta}_k^\ast \} \big]_{ij} \big| \ge C_{0q} \sqrt{\frac{\ln(M^{1/\tau} q)}{Kp}} \; \Big) \nonumber \\
		& \quad \le \frac{1}{q^{\tau -2}} + 16 M q^2 e^{- Kp/2}     \label{apeqn360}
\end{align}
for any $q \ge 1$, where $C_{0q}$ is given by (\ref{eqn410}) and
\begin{align}
	E\{ \tilde{\bm \Theta}_k^\ast \} & = \frac{1}{p} \, \mbox{tr}\big(\hat{\bm \Omega} {\bm \Sigma}^\diamond \big) \, 
	    (\bar{\bm S}^\diamond_k)^\ast \, . \label{apeqn363}
\end{align}
{\it Proof}.  Let ${\bm D}_z(\tilde{f}_{k,\ell}) = {\bm D}_{r,kl} + j {\bm D}_{i,kl}$, ${\bm D}_{r,kl}, {\bm D}_{i,kl} \in \mathbb{R}^{p \times q}$. Define
\begin{align}
  {\bm X}_{kl} = & \begin{bmatrix} {\bm D}_{r,kl} & {\bm D}_{i,kl} \end{bmatrix} \in \mathbb{R}^{p \times (2q)} \, , \label{apeqn365} \\
	{\bm B}_{kl} = & {\bm X}_{kl}^\top \hat{\bm \Omega} {\bm X}_{kl} \, , \quad
	   {\bm F}_{k} =  \frac{1}{Kp} \sum_{\ell = -m_t}^{m_t} {\bm B}_{kl} \, . \label{apeqn366}
\end{align}
Since
\begin{align}
  {\bm D}_z^H &(\tilde{f}_{k,\ell}) \hat{\bm \Omega} {\bm D}_z(\tilde{f}_{k,\ell}) 
	  = {\bm D}_{r,kl}^\top \hat{\bm \Omega} {\bm D}_{r,kl}  
		  + {\bm D}_{i,kl}^\top \hat{\bm \Omega} {\bm D}_{i,kl}   \nonumber \\
		& \;\; 	+j \big[ {\bm D}_{r,kl}^\top \hat{\bm \Omega} {\bm D}_{i,kl} 
		  - {\bm D}_{i,kl}^\top \hat{\bm \Omega} {\bm D}_{r,kl} \big] \, , \label{apeqn368} 
\end{align}
it follows that
\begin{align}
 \max_{k,i,j} & \big| \big[ \tilde{\bm \Theta}_k^\ast -
	       E\{ \tilde{\bm \Theta}_k^\ast \} \big]_{ij} \big|  \le
				4 \, \max_{k,i,j} \big| \big[ {\bm F}_{k} -
	       E\{ {\bm F}_{k} \} \big]_{ij} \big| \, . \label{apeqn370} 
\end{align}
Therefore,
\begin{align}
 \Big\{ \max_{k,i,j} & \big| \big[ {\bm F}_{k} -
	       E\{ {\bm F}_{k} \} \big]_{ij} \big|  < \frac{\delta}{4} \Big\} \nonumber \\ 
				& \quad \subseteq 
 \Big\{ \max_{k,i,j}  \big| \big[ \tilde{\bm \Theta}_k^\ast -
	       E\{ \tilde{\bm \Theta}_k^\ast \} \big]_{ij} \big|  < \delta \Big\} \, ,  \label{apeqn371} 
\end{align}
implying
\begin{align}
  P & \Big( \max_{k,i,j}  \big| \big[ \tilde{\bm \Theta}_k^\ast -
	       E\{ \tilde{\bm \Theta}_k^\ast \} \big]_{ij} \big| \ge \delta \Big) \nonumber \\
		& \le P  \Big( \max_{k,i,j}  \big| \big[ {\bm F}_{k} -
	       E\{ {\bm F}_{k} \} \big]_{ij} \big|  \ge \frac{\delta}{4}  \Big) \, .  \label{apeqn372}
\end{align}
Since ${\bm d}_z(\tilde{f}_{k,\ell}) = \mbox{vec}({\bm D}_z(\tilde{f}_{k,\ell})) \sim {\mathcal N}_c( {\bf 0}, \bar{\bm S}^\diamond(\tilde{f}_k) \otimes {\bm \Sigma}^\diamond)$, it follows from Lemma 3(i) that 
\begin{align}
  {\bm X}_{kl} \sim & \mathcal{MVN} ({\bm 0}, {\bm \Sigma}^\diamond, \tilde{\bm S}^\diamond_k ) \, , \label{apeqn375} \\
	\tilde{\bm S}^\diamond_k =	& \frac{1}{2} \begin{bmatrix}
	    \bar{\bm S}^\diamond_{rk} & -\bar{\bm S}^\diamond_{ik} \\ \bar{\bm S}^\diamond_{ik} & \bar{\bm S}^\diamond_{rk} \end{bmatrix} \, ,
				\quad \bar{\bm S}^\diamond(\tilde{f}_k) = \bar{\bm S}^\diamond_{rk} + j \bar{\bm S}^\diamond_{ik} \, . \label{apeqn377}
\end{align}
By assumption (A4), $\phi_{\max}({\bm \Sigma}^\diamond)  \le \beta_{p,\max}$ and additionally, by Lemma 4, $\phi_{\max}(\tilde{\bm S}^\diamond_k)  \le \beta_{q,\max}/2$ for every $k$. With $a=4(1+\gamma_p \beta_{p,\max})\beta_{q,\max}$, invoking Lemma 1 for the sum ${\bm F}_k$, we have
\begin{align}
& P \Big( \max_{i,j}  \big| \big[ {\bm F}_{k} -
	       E\{ {\bm F}_{k} \} \big]_{ij} \big|  \ge \frac{\delta}{4}  \Big) \nonumber \\
	& \le 4 (2q) ^2 \Big[ \exp\big\{ - \frac{Kp}{2} \big[ \frac{\delta/4}{a} 
	        - \frac{2}{\sqrt{Kp}}  \big]^2 \big\}   + e^{-Kp/2 } \Big]  = P_{qtb} \, .  \label{apeqn380}
\end{align}
Maximizing over all $k=1,2, \cdots , M$, and using the union bound, we obtain
\begin{align}
 P & \Big( \max_{k,i,j}  \big| \big[ {\bm F}_{k} -
	       E\{ {\bm F}_{k} \} \big]_{ij} \big|  \ge \frac{\delta}{4}  \Big) \le M P_{qtb} \, .  \label{apeqn381}
\end{align}
For $\tau > 2$, pick $\delta = 4a ( \sqrt{2 \, \ln(16 M q^\tau )/(Kp)} + 2/\sqrt{Kp} \,)$, leading to $\delta = C_{0q} \sqrt{\ln(M^{1/\tau}q)/(Kp)}$ and $(Kp/2)[(\delta/(4a))-2/\sqrt{Kp}]^2= \ln(16 M \, q^\tau)$. Thus
\begin{align}
M P_{qtb} = & 16 M q^2 \Big[ e^{-\ln(16 M q^\tau)}   + e^{-Kp/2 } \Big] \nonumber \\
 = & \frac{1}{q^{\tau -2}} + 16 M q^2 e^{-Kp/2 } 
  \, .  \label{apeqn382}
\end{align}
Thus we have established (\ref{apeqn360}). The lower bound on $\delta/4$ specified in Lemma 1 is satisfied if $(\delta/(4a) > 2/\sqrt{Kp}$, which is true for any $q \ge 1$. Turning to (\ref{apeqn363}), by (\ref{apeqn366}), (\ref{apeqn375}) and Lemma 2(iii), we have
\begin{align}
E\{ {\bm B}_{kl} \} = & \mbox{tr}( \hat{\bm \Omega} {\bm \Sigma}^\diamond ) \frac{1}{2} \begin{bmatrix}
	    \bar{\bm S}^\diamond_{rk} & -\bar{\bm S}^\diamond_{ik} \\ \bar{\bm S}^\diamond_{ik} & \bar{\bm S}^\diamond_{rk} \end{bmatrix}
  \, .  \label{apeqn385}
\end{align}
By assumption (A1), (\ref{apeqn366}), (\ref{apeqn368}) and (\ref{apeqn385}), 
\begin{align}
  E \{ {\bm D}_z^H &(\tilde{f}_{k,\ell}) \hat{\bm \Omega} {\bm D}_z(\tilde{f}_{k,\ell}) \}
	  = \frac{1}{2} \mbox{tr}( \hat{\bm \Omega} {\bm \Sigma}^\diamond ) 
		\Big( 2 \bar{\bm S}^\diamond_{rk}  - j 2 \bar{\bm S}^\diamond_{ik}\Big) \nonumber \\
		& = \mbox{tr}( \hat{\bm \Omega} {\bm \Sigma}^\diamond ) (\bar{\bm S}^\diamond_k)^\ast \, . \label{apeqn386} 
\end{align}
By (\ref{eqn252}) and (\ref{apeqn386}), we obtain  (\ref{apeqn363}). 
$\quad \Box$

Now we consider a tail bound on $\check{\bm \Theta} $ defined in (\ref{eqn242}). \\
{\it Lemma 6}. Under assumptions (A1) and (A2), for any Hermitian positive-definite $\hat{\bm \Phi}_k \in \mathbb{C}^{q \times q}$, $k=1,2, \cdots , M$,  such that $\| \hat{\bm \Gamma} - {\bm \Gamma}^\diamond\|_F \le \gamma_q$, ${\bm \Gamma}^\diamond =[{\bm \Phi}_1^\diamond, \; \cdots , {\bm \Phi}_M^\diamond]$, $\hat{\bm \Gamma} =[\hat{\bm \Phi}_1, \; \cdots , \hat{\bm \Phi}_M]$,  and $\tau >2$, we have
\begin{align}
  P & \Big( \max_{i,j} \big| \big[ \check{\bm \Theta} -
	       E\{ \check{\bm \Theta} \} \big]_{ij} \big| \ge C_{0p} \sqrt{\frac{\ln(p)}{KqM}} \, \Big) \nonumber \\
		& \quad \le  \frac{1}{p^{\tau-2}} + 4 p^2 e^{ - KqM}   \label{apeqn390}
\end{align}
for any $p \ge 1$ where, where $C_{0p}$ is given by (\ref{eqn412}), and
\begin{align}
	E\{ \check{\bm \Theta} \} & = \Big[ \frac{1}{2Mq} \sum_{k=1}^M \mbox{tr} \big( \bar{\bm S}^\diamond_k \hat{\bm \Phi}_k 
	  + ( \bar{\bm S}^\diamond_k \hat{\bm \Phi}_k )^\ast \big)
	     \Big] \, {\bm \Sigma}^\diamond \, . \label{apeqn393}
\end{align}
{\it Proof}.  We have
\begin{align}
  \mbox{Re}\big( {\bm D}_z &(\tilde{f}_{k,\ell}) \hat{\bm \Phi}_k {\bm D}_z^H(\tilde{f}_{k,\ell}) \big)
	  = {\bm X}_{kl} \tilde{\bm \Phi}_k {\bm X}_{kl}^\top \, , \label{apeqn400} 
\end{align}
where ${\bm X}_{kl}$ is as in (\ref{apeqn365}) and 
\begin{equation} \label{apeqn403}
  \tilde{\bm \Phi}_k = \begin{bmatrix}
	    \hat{\bm \Phi}_{rk} & -\hat{\bm \Phi}_{ik} \\ \hat{\bm \Phi}_{ik} & \hat{\bm \Phi}_{rk} 
			   \end{bmatrix} \in \mathbb{R}^{2q \times 2q} \; .
\end{equation}
Define
\begin{align} 
  \check{\bm \Phi} = & \begin{bmatrix}
	    \tilde{\bm \Phi}_{1} & {\bm 0} & \cdots & {\bm 0} \\ 
			{\bm 0} & \tilde{\bm \Phi}_{2} & \cdots &  {\bm 0} \\
			\vdots & \vdots  & \ddots & \vdots \\
			{\bm 0} & {\bm 0} & \cdots & \tilde{\bm \Phi}_{M}
			   \end{bmatrix} \in \mathbb{R}^{(2qM) \times (2qM)} \; ,  \label{apeqn405} \\
	\check{\bm X}_\ell = & \begin{bmatrix} {\bm X}_{1l} & {\bm X}_{2l} 
	          & \cdots & {\bm X}_{Ml} \end{bmatrix}^\top \; .  \label{apeqn406}
\end{align}
Then we can express $\check{\bm \Theta}$ as
\begin{align} 
  \check{\bm \Theta} = & \frac{1}{MKq} \sum_{\ell=-m_t}^{m_t}  
	  \check{\bm X}_\ell^\top \check{\bm \Phi} \check{\bm X}_\ell  \; .  \label{apeqn407}
\end{align}
Since ${\bm X}_{kl} \sim  \mathcal{MVN} ({\bm 0}, {\bm \Sigma}^\diamond, \tilde{\bm S}^\diamond_k )$, and ${\bm X}_{k_1 l}$ and ${\bm X}_{k_2 l}$ are independent for $k_1 \ne k_2$, we have
\begin{align} 
  \check{\bm X}_\ell^\top \sim & \mathcal{MVN} ({\bm 0}, {\bm \Sigma}^\diamond, \check{\bm S}^\diamond ) \, ,\;
	\check{\bm X}_\ell \sim  \mathcal{MVN} ({\bm 0}, \check{\bm S}^\diamond, {\bm \Sigma}^\diamond ) \, , \label{apeqn409}
\end{align}
where
\begin{align} 
  \check{\bm S}^\diamond = & \begin{bmatrix}
	    \tilde{\bm S}^\diamond_{1} & {\bm 0} & \cdots & {\bm 0} \\ 
			{\bm 0} & \tilde{\bm S}^\diamond_{2} & \cdots &  {\bm 0} \\
			\vdots & \vdots  & \ddots & \vdots \\
			{\bm 0} & {\bm 0} & \cdots & \tilde{\bm S}^\diamond_{M}
			   \end{bmatrix} \in \mathbb{R}^{(2qM) \times (2qM)} \; .  \label{apeqn410}
\end{align}
By assumption (A4), $\phi_{\max}({\bm \Sigma}^\diamond)  \le \beta_{p,\max}$ and additionally, by Lemma 4, $\phi_{\max}(\check{\bm S}^\diamond)  \le \beta_{q,\max}/2$. With $b=8(1+\gamma_q \beta_{q,\max}/2)\beta_{p,\max}$, apply Lemma 1 to the sum $\frac{1}{2} \check{\bm \Theta}$ to obtain
\begin{align}
 P &\Big( \max_{i,j}  \big| \frac{1}{2} \big[ \check{\bm \Theta} -
	       E\{ \check{\bm \Theta} \} \big]_{ij} \big|  \ge \delta \Big) \nonumber \\
	& \le 4 p^2 \Big[ \exp\big\{ - \frac{2qMK}{2} \big[ \frac{\delta}{b} 
	        - \frac{2}{\sqrt{2qMK}}  \big]^2 \big\}   + e^{-2qMK/2 } \Big] \nonumber \\
	& =  P_{ptb} \, .  \label{apeqn420}
\end{align}
For $\tau > 2$, pick $\delta = b ( \sqrt{\ln(4 p^\tau)/(KqM)} + \sqrt{2/(KqM)} \,)$, leading to $(2qMK/2)[(\delta/b)-\sqrt{2/(KqM)}]^2= \ln(4 p^\tau)$. Thus
\begin{align}
P_{ptb} = & 4 p^2 \Big[ e^{-\ln(4 p^\tau)}   + e^{-qMK } \Big]  =  \frac{1}{p^{\tau-2}} + 4 p^2 e^{-qMK }
  \, .  \label{apeqn422}
\end{align}
The lower bound on $\delta$ specified in Lemma 1 is satisfied if $ (\delta/b) > \sqrt{2/(KqM)}$, which is true for any $p \ge 1$. With our choice of $\delta$, we have $2 \delta = C_{0p} \sqrt{\frac{\ln(p)}{KqM}}$, establishing (\ref{apeqn390}). Turning to (\ref{apeqn393}), by (\ref{apeqn409}) and Lemma 2(iii), we have
\begin{align}
  E & \{ \check{\bm X}_\ell^\top \check{\bm \Phi} \check{\bm X}_\ell \} 
	 = \mbox{tr} \big( \check{\bm \Phi}^\top \check{\bm S}^\diamond \big) \, {\bm \Sigma}^\diamond
	 = \mbox{tr} \big( \sum_{k=1}^M \tilde{\bm \Phi}_k^\top \tilde{\bm S}^\diamond_k \big) \, {\bm \Sigma}^\diamond \, .
	   \label{apeqn424}
\end{align}
By (\ref{apeqn377}) and (\ref{apeqn403}), 
\begin{align}
 \mbox{tr}  \big( \tilde{\bm \Phi}_k^\top \tilde{\bm S}^\diamond_k \big)
   & = \mbox{tr} \big( \hat{\bm \Phi}_{rk} \bar{\bm S}_{rk}^\diamond + \hat{\bm \Phi}_{ik} \bar{\bm S}_{ik}^\diamond \big) \nonumber \\
   &  = \frac{1}{2} \, \mbox{tr} \big( \bar{\bm S}^\diamond_k \hat{\bm \Phi}_k 
	  + ( \bar{\bm S}^\diamond_k \hat{\bm \Phi}_k )^\ast \big)   \, .          \label{apeqn425}
\end{align}
Using (\ref{apeqn407}), (\ref{apeqn424}) and (\ref{apeqn425}), we have (\ref{apeqn393}). $\quad \Box$

{\it Proof of Theorem 2(i)}. Let ${\bm \Omega} = \bar{\bm \Omega}({\bm \Gamma}) + {\bm \Delta}$ with ${\bm \Omega}, \bar{\bm \Omega}({\bm \Gamma}) \succ {\bm 0}$, and denote $Q({\bm \Omega}) = L_1({\bm \Omega}) - L_1(\bar{\bm \Omega}({\bm \Gamma}))$. For the rest of the proof, we will denote $\bar{\bm \Omega}({\bm \Gamma})$ by $\bar{\bm \Omega}$. Then $\hat{\bm \Omega}({\bm \Gamma})$ minimizes $Q({\bm \Omega})$, or equivalently, $\hat{\bm \Delta} = \hat{\bm \Omega}({\bm \Gamma}) - \bar{\bm \Omega}$ minimizes $J({\bm \Delta}) = Q(\bar{\bm \Omega} + {\bm \Delta})$. Consider the set
\begin{equation}  \label{apeqn500}
  \Psi_p(R_p) :=  \left\{ \bm{\Delta} \, :\, \bm{\Delta} = \bm{\Delta}^\top \;, \; \|\bm{\Delta} \|_F = R_p r_{pn} \right\}
\end{equation}
where $R_p = 17 C_{0p}/\beta_{p,\min}^2$ and $r_{pn}$ is as in (\ref{eqn417}). Since $J(\hat{\bm \Delta}) \le J({\bm 0}) = 0$, if we can show that $\inf_{\bm{\Delta}}  \{ J(\bm{\Delta}) \, :\, \bm{\Delta} \in \Psi_p(R_p) \} \, > \, 0 $, then the minimizer $\hat{\bm{\Delta}}$ must be inside the sphere defined by $\Psi_p(R_p)$, and hence, $\| \hat{\bm{\Delta}} \|_F \le   R_p r_{pn}$. It is shown in \cite[(9)]{Rothman2008} that $\ln (|\bar{\bm \Omega} +{\bm  \Delta}|) - \ln (|\bar{\bm \Omega}|) = \mbox{tr} (\bar{\bm \Omega}^{-1}  {\bm \Delta}) - \tilde{B}_1$ 
where, with $\bm{H}(\bar{\bm \Omega}, \bm{\Delta}, v ) = (\bar{\bm \Omega} + v \bm{\Delta})^{-1} \otimes (\bar{\bm \Omega}+v \bm{\Delta})^{-1}$ and $v$ denoting a real scalar,
\begin{align} \label{apeqn502}
     \tilde{B}_1 := & \mbox{vec}(\bm{\Delta})^\top \left( \int_0^1 (1-v) 
			  \bm{H}(\bar{\bm \Omega}, \bm{\Delta}, v ) \, dv \right)  \mbox{vec}(\bm{\Delta}) \, .
\end{align} 
We have
\begin{align} 
  J(\bm{\Delta}) = & \sum_{i=1}^3 B_i \, , \quad B_1 = \frac{1}{p} \tilde{B}_1 \, , \label{apeqn504} \\
     B_2  := & \frac{1}{p} \mbox{tr} \big( (\check{\bm{\Theta}} - \bar{\bm \Omega}^{-1} ) \bm{\Delta}  \big) \, ,  \label{apeqn505} \\
	   B_3  := & \lambda_p \big( \| \bar{\bm \Omega}^- + {\bm \Delta}^- \|_1 - \| \bar{\bm \Omega}^- \|_1 \big) \, .  \label{apeqn506}
\end{align}
By (\ref{aeqn254}) and (\ref{apeqn393}), $\bar{\bm \Omega}^{-1} = E\{ \check{\bm \Theta} \}  = \Big( \frac{1}{2Mq} \sum_{k=1}^M \mbox{tr} \big( \bar{\bm S}^\diamond_k {\bm \Phi}_k  + ( \bar{\bm S}^\diamond_k {\bm \Phi}_k )^\ast \big) \Big) \, {\bm \Sigma}^\diamond$ (where we replaced $\hat{\bm \Phi}_k$ with ${\bm \Phi}_k$). By Lemma 6, $\max_{i,j} \big| \big[ \check{\bm \Theta} - E\{ \check{\bm \Theta} \} \big]_{ij} \big| \ge C_{0p} \sqrt{\frac{\ln(p)}{KqM}}$ w.h.p.\ (which refers to  $1-\frac{1}{p^{\tau-2}} - 4 p^2 e^{ - KqM}$, cf.\ (\ref{apeqn390})). Following \cite[p.\ 502]{Rothman2008}, we have
\begin{equation}
     \tilde{B}_1  \ge  \| \bm{\Delta} \|_F^2 / \big(  2 (\| \bar{\bm \Omega} \| +  \| \bm{\Delta} \|)^2 \big) \, .  \label{apeqn508}
\end{equation}
Turning to $E\{ \check{\bm \Theta} \} $, we have
\begin{align} 
  \mbox{tr} & \big( \bar{\bm S}^\diamond_k {\bm \Phi}_k + ( \bar{\bm S}^\diamond_k {\bm \Phi}_k )^\ast \big) 
	  = 2 \mbox{Re} \, \mbox{tr}  \big( \bar{\bm S}^\diamond_k ({\bm \Phi}_k - {\bm \Phi}_k^\diamond + {\bm \Phi}_k^\diamond ) \big) 
		   \label{apeqn510} \\
     & = 2 \mbox{Re} \, \mbox{tr}  \big( \bar{\bm S}^\diamond_k ({\bm \Phi}_k - {\bm \Phi}_k^\diamond ) \big) 
		   + 2 \mbox{tr}({\bm I}_q)   \label{apeqn511} \\
		&  \ge 2 q - 2 \, \| \bar{\bm S}^\diamond_k \|_F \, \| {\bm \Phi}_k - {\bm \Phi}_k^\diamond \|_F   \label{apeqn512}
\end{align}
where we used $|\mbox{tr}({\bm B} {\bm C}^H)| \le \| {\bm B}\|_F \, \| {\bm C}\|_F $ (Cauchy-Schwarz inequality). Since $\| \bar{\bm S}^\diamond_k \|_F \le \sqrt{q} \, \| \bar{\bm S}^\diamond_k \| \le \sqrt{q} \, \beta_{q,\max}$ and $\sum_{k=1}^M \| {\bm \Phi}_k - {\bm \Phi}_k^\diamond \|_F \le \sqrt{M} \| {\bm \Gamma} - {\bm \Gamma}^\diamond \|_F \le \sqrt{M} \, \gamma_q$, we have
\begin{align} 
  A & =  2 \mbox{Re} \, \sum_{k=1}^M \mbox{tr}  \big( \bar{\bm S}^\diamond_k {\bm \Phi}_k \big)   
		\ge 2 M q - 2 \sqrt{Mq} \, \beta_{q,\max}  \gamma_q \label{apeqn514} \\
		&  \ge 2 M q - 2 M q \, \beta_{q,\max}  \gamma_q  = 1.8 M q \, , \label{apeqn515}
\end{align}
where we have used the facts that $\sqrt{Mq} \le Mq$ and $\beta_{q,\max} \gamma_q = 0.1$, as defined in (\ref{eqn410c}). Therefore, $\| \bar{\bm \Omega}^{-1} \| = \| E\{ \check{\bm \Theta} \} \|  \ge 0.9 \, \| {\bm \Sigma}^\diamond \|$, implying $\| \bar{\bm \Omega} \| \le 10/(9 \, \| {\bm \Sigma}^\diamond \|) \le 10/(9 \, \beta_{p,\min}) \le 1.5/\beta_{p,\min} \,$. Using (\ref{apeqn504}), (\ref{apeqn508}), and the facts  $\| \bar{\bm \Omega} \| \le 1.5/\beta_{p,\min}$ and $\| \bm{\Delta} \| \le \| \bm{\Delta} \|_F = R_p r_{pn}$, we obtain w.h.p.
\begin{equation}
     {B}_1  \ge  \| \bm{\Delta} \|_F^2 \, \beta_{p,\min}^2 / (  8 p ) \, ,  \label{apeqn518}
\end{equation}
for $n > N_p$, since $r_{pn} \le \beta_{p,\min}/(34 C_{0p})$ for $n > N_p$ and $R_p r_{pn} \le 0.5/\beta_{p,\min}$.

We now consider $B_2$ given by (\ref{apeqn505}). Define $\bar{\cal S}_p = {\cal S}_p \cup \{ \{i,j\}: \, i=j \}$ so that $|\bar{\cal S}_p| = s_p+p$. We have
\begin{align*}
     |B_2|  & \le  B_{12}+B_{22} , \;\; pB_{12} = \Big| \sum_{\{i,j\} \in \bar{\cal S}_p } 
			   [\check{\bm{\Theta}} - \bar{\bm \Omega}^{-1}]_{ij} \Delta_{ji} \, \Big| \, , \\
		pB_{22} & = \Big| \sum_{\{i,j\} \in \bar{\cal S}_p^c } 
			   [\check{\bm{\Theta}} - \bar{\bm \Omega}^{-1}]_{ij} \, \Delta_{ji} \, \Big|  \, ,
\end{align*}
where $\bar{\cal S}_p^c$ denotes the complement of set $\bar{\cal S}_p$.
For an index set ${\bm B}$ and a matrix ${\bm C} \in \mathbb{R}^{p \times p}$, we write ${\bm C}_{\bm B}$ to denote a matrix in $\mathbb{R}^{p \times p}$ such that $[{\bm C}_{\bm B}]_{ij} = C_{ij}$ if $(i,j) \in {\bm B}$, and $[{\bm C}_{\bm B}]_{ij}=0$ if $(i,j) \not\in {\bm B}$. Using $\big| \sum_{\{i,j\} \in \bar{\cal S}_p }  \Delta_{ij} \big| \le \sqrt{s_p+p} \, \| {\bm \Delta} \|_F$ (by Cauchy-Schwarz inequality), 
\begin{align}
 p&B_{12}  \le  \max_{i,j} \, [\check{\bm{\Theta}} - \bar{\bm \Omega}^{-1}]_{ij} \;
       \big| \sum_{\{i,j\} \in \bar{\cal S}_p }  \Delta_{ij} \big| \, , \nonumber \\
		& \le C_{0p} \sqrt{\ln(p)/(KqM)} \, \sqrt{s_p+p} \, \| {\bm \Delta} \|_F  = C_{0p} r_{pn} \| {\bm \Delta} \|_F \, . \label{apeqn521}
\end{align}
We will combine $B_{22}$ with $B_3$. By (\ref{apeqn506}),
\begin{align} 
	   B_3  = & \lambda_p \big( \| \bar{\bm \Omega}^- + {\bm \Delta}_{{\cal S}_p}^- \|_1 
		   + \| {\bm \Delta}_{{\cal S}_p^c}^- \|_1 - \| \bar{\bm \Omega}^- \|_1 \big) \nonumber \\
			\ge &  \lambda_p \big( \| {\bm \Delta}_{{\cal S}_p^c}^- \|_1 - \|  {\bm \Delta}_{{\cal S}_p}^- \|_1  \big) \, ,  \label{apeqn523}
\end{align}
using the triangle inequality $\| \bar{\bm \Omega}^- + {\bm \Delta}_{{\cal S}_p}^- \|_1 \ge \| \bar{\bm \Omega}^-  \|_1 - \|  {\bm \Delta}_{{\cal S}_p}^- \|_1$ and the fact $\bar{\bm \Omega}_{{\cal S}_p^c}^- = \bar{\bm \Omega}_{\bar{\cal S}_p^c} = {\bm 0}$. Hence, $B_2+B_3 \ge -B_{12} - B_{22} + \lambda_p \big( \| {\bm \Delta}_{{\cal S}_p^c}^- \|_1 - \|  {\bm \Delta}_{{\cal S}_p}^- \|_1  \big)$. But $p B_{22} \le C_{0p} \sqrt{\ln(p)/(KqM)} \, \| {\bm \Delta}_{{\cal S}_p^c}^- \|_1$ w.h.p., therefore,
\begin{align} 
	  B_2 + B_3  \ge & \big( \lambda_p - C_{0p} \sqrt{\ln(p)/(p^2 KqM)} \, \big) \| {\bm \Delta}_{{\cal S}_p^c}^- \|_1 \nonumber \\
		   & - \lambda_p \|  {\bm \Delta}_{{\cal S}_p}^- \|_1 
			 - C_{0p} r_{pn} \| {\bm \Delta} \|_F /p \, .  \label{apeqn525}
\end{align}
Using the fact that by (\ref{eqn452}), the first term on right side of (\ref{apeqn525}) is nonnegative, and $\|  {\bm \Delta}_{{\cal S}_p}^- \|_1 \le \sqrt{s_p} \, \| {\bm \Delta} \|_F$ by the Cauchy-Schwarz inequality, we obtain $B_2 + B_3  \ge - \big(\lambda_p \sqrt{s_p} +  r_{pn}/p \big) \| {\bm \Delta} \|_F $. Thus, by (\ref{eqn452}), (\ref{apeqn504}) and (\ref{apeqn518}) 
\begin{align} 
	  J({\bm \Delta})  \ge & \frac{ \| \bm{\Delta} \|_F^2 \, \beta_{p,\min}^2}{  8 p } -
		  \big(\lambda_p \sqrt{s_p} + C_{0p} r_{pn} /p \big) \| {\bm \Delta} \|_F  \nonumber \\
			\ge & \frac{ \| \bm{\Delta} \|_F^2 \, \beta_{p,\min}^2}{  8 p } -
		  \frac{2 C_{0p} r_{pn}  \| {\bm \Delta} \|_F}{p} \nonumber \\
		  = & 
			 \frac{ \| \bm{\Delta} \|_F^2 \, \beta_{p,\min}^2}{  8 p } \Big( 1- \frac{16}{17} \Big) > 0  \label{apeqn530}
\end{align}
using $\| {\bm \Delta} \|_F = R_p r_{pn}$ and $R_p = 17 C_{0p}/\beta_{p,\min}^2$. This proves Theorem 2(i). $\quad \Box$ 

{\it Proof of Theorem 2(ii)}. With ${\bm \Gamma}$ as in (\ref{eqn400}), let ${\bm \Gamma} = \bar{\bm \Gamma}({\bm \Omega}) + {\bm \Lambda}$ with ${\bm \Phi}_k = {\bm \Phi}_k ^H \succ {\bm 0}$, and denote $Q({\bm \Gamma}) = L_2({\bm \Gamma}) - L_2(\bar{\bm \Gamma}({\bm \Omega}))$. For the rest of the proof, we will denote $\bar{\bm \Gamma}({\bm \Omega})$ by $\bar{\bm \Gamma}$. Then $\hat{\bm \Gamma}({\bm \Omega})$ minimizes $Q({\bm \Gamma})$, or equivalently, $\hat{\bm \Lambda} = \hat{\bm \Gamma}({\bm \Omega}) - \bar{\bm \Gamma}$ minimizes $J({\bm \Lambda}) = Q(\bar{\bm \Gamma} + {\bm \Lambda})$. Note that ${\bm \Lambda} = [{\bm \Lambda}_1 , \; \cdots , \; {\bm \Lambda}_M] \in \mathbb{C}^{q \times (qM)}$ and ${\bm \Lambda}_k = {\bm \Phi}_k - \bar{\bm \Phi}_k$, $k=1, \cdots , M$, where $\bar{\bm \Phi}_k = \bar{\bm \Phi}_k({\bm \Omega}) =  p \bm{\Phi}_k^\diamond / \mbox{tr}( {\bm \Sigma}^\diamond {\bm \Omega} )$ by (\ref{aeqn256}). Consider the set
\begin{equation}  \label{apeqn600}
  \Psi_q(R_q) :=  \left\{ \bm{\Lambda} \, :\, \bm{\Lambda}_k = \bm{\Lambda}_k^H, \, k=1, \cdots M, \, \|\bm{\Lambda} \|_F = R_q r_{qn} \right\}
\end{equation}
where $R_q = 17 C_{0q}/\beta_{q,\min}^2$ and $r_{qn}$ is as in (\ref{eqn415}). Similar to the proof of Theorem 2(i), our objective is to show   that $\inf_{\bm{\Lambda}}  \{ J(\bm{\Lambda}) \, :\, \bm{\Lambda} \in \Psi_q(R_q) \} \, > \, 0 $, which would ensure $\| \hat{\bm{\Lambda}} \|_F \le   R_q r_{qn}$ w.h.p. It is shown in \cite[Lemma 5]{Tugnait22c} that $\ln (|\bar{\bm \Phi}_k +{\bm \Lambda}_k|) -\ln (|\bm{\Phi}_k|) +\ln ( |\bar{\bm \Phi}_k^\ast +{\bm \Lambda}_k^\ast| ) -\ln (|\bm{\Phi}_k^\ast|) = \mbox{tr} \big(\bar{\bm \Phi}_k^{-1}  {\bm \Lambda}_k + (\bar{\bm \Phi}_k^{-1}  {\bm \Lambda}_k)^\ast \big) - \tilde{B}_{1k}$ 
where 
\begin{align} 
     \tilde{B}_{1k} = & \bm{g}^H(\bm{\Lambda}_k) \left( \int_0^1 (1-v) 
			  \bm{H}_k(\bar{\bm \Phi}_k, \bm{\Lambda}_k, v ) \, dv \right)  \bm{g}(\bm{\Lambda}_k) \, , \label{apeqn602} \\
		\bm{g}(\bm{\Lambda}_k) = & \begin{bmatrix} \mbox{vec}(\bm{\Lambda}_k) \\
	       \mbox{vec}(\bm{\Lambda}_k^\ast)  \end{bmatrix} , \;
  \bm{H}_k(\bar{\bm \Phi}_k, \bm{\Lambda}_k, v ) 
	    = \begin{bmatrix} \bm{H}_{11k} &  \bm{0}  \\
	       \bm{0} & \bm{H}_{22k}  \end{bmatrix} \, , \label{apeqn603} \\
	\bm{H}_{11k} = & (\bar{\bm \Phi}_k +v \bm{\Lambda}_k)^{-\ast} \otimes (\bar{\bm \Phi}_k +v \bm{\Lambda}_k)^{-1} \, , \label{apeqn604}
\end{align}
\begin{align} 
	\bm{H}_{22k} = & (\bar{\bm \Phi}_k +v \bm{\Lambda}_k)^{-1} \otimes (\bar{\bm \Phi}_k +v \bm{\Lambda}_k)^{-\ast} \, , \label{apeqn605}
\end{align}
and $v$ is a real scalar. Therefore,
\begin{align} 
  J(\bm{\Lambda}) = & \sum_{k=1}^M \sum_{i=1}^3 B_{ik} + B_4 \, , \quad B_{ik} = \frac{1}{2Mq} \tilde{B}_{1k} \, , \label{apeqn610} \\
     B_{2k}  = & \frac{1}{2Mq} \mbox{tr} \big( \tilde{B}_{2k} + \tilde{B}_{2k}^\ast  \big) \, , 
			  \quad \tilde{B}_{2k} = (\tilde{\bm{\Theta}} - \bar{\bm \Phi}_k^{-1} ) \bm{\Lambda}_k \, , \label{apeqn611} \\
	   B_{3k}  = & \alpha \lambda_q \big( \| \bar{\bm \Phi}_k^- + {\bm \Lambda}_k^- \|_1 
		         - \| \bar{\bm \Phi}_k^- \|_1 \big) \, ,  \label{apeqn612} \\
		B_4 = & (1-\alpha) \sqrt{M} \lambda_q \, \sum_{ i \ne j}^p \; \big( \| {\bm{\Phi}}^{(ij)} + {\bm{\Lambda}}^{(ij)} \|
		   - \| {\bm{\Phi}}^{(ij)} \| \big) \, . \label{apeqn613}
\end{align}
By \cite[Eqn.\ (B.43)]{Tugnait22c}, we have
\begin{equation}
     B_{1k}  \ge \frac{1}{2Mq} \, \frac{ \| \bm{\Lambda}_k \|_F^2 }{   (\| \bar{\bm \Phi}_k \| 
		           +  \| \bm{\Lambda}_k \|)^2 } \, .  \label{apeqn620}
\end{equation}
Now $\mbox{tr}( {\bm \Sigma}^\diamond {\bm \Omega} ) = \mbox{tr}( {\bm \Sigma}^\diamond ({\bm \Omega} -{\bm \Omega}^\diamond + {\bm \Omega}^\diamond)) = \mbox{tr}( {\bm \Sigma}^\diamond ({\bm \Omega} -{\bm \Omega}^\diamond) ) +p$. Since $|\mbox{tr}( {\bm \Sigma}^\diamond ({\bm \Omega} -{\bm \Omega}^\diamond) )| \le \|{\bm \Sigma}^\diamond\|_F \,  \|{\bm \Omega} -{\bm \Omega}^\diamond \|_F \le \sqrt{p} \, \beta_{p,\max} \gamma_p$, we have $|\mbox{tr}( {\bm \Sigma}^\diamond {\bm \Omega} )| \ge p -  \sqrt{p} \, \beta_{p,\min} \gamma_p \ge p -  p \, \beta_{p,\min} \gamma_p = 0.9 p$ since $\gamma_p = 0.1/\beta_{p,\min}$. Therefore, $\|\bar{\bm \Phi}_k\| \le  p \| \bm{\Phi}_k^\diamond \| / (0.9 p) \le 1.5/\beta_{q,\min}$. Also, $\|{\bm \Lambda}_k\| \le \|{\bm \Lambda}_k\|_F \le \|{\bm \Lambda}\|_F = R_q r_{qn}$. Therefore,
\begin{align}
    \sum_{k=1}^M B_{1k}  \ge & \frac{1}{2Mq} \, \frac{ \sum_{k=1}^M\| \bm{\Lambda}_k \|_F^2 }{   (1.5/\beta_{q,\min} 
		           +  R_q r_{qn})^2 } \nonumber \\
				\ge & \frac{ \| \bm{\Lambda} \|_F^2 \, \beta_{q,\min}^2}{   8 M q}    \label{apeqn622}
\end{align}
w.h.p.\ for $n > N_q$, since $r_{qn} \le \beta_{q,\min}/(34 C_{0q})$ for $n > N_q$ and $R_q r_{qn} \le 0.5/\beta_{q,\min}$.

We now bound $B_{2k}$ noting that $|B_{2k}|   \le  L_{1k} + L_{2k}$ where
\begin{align*}
      L_{1k} & = \frac{2}{2 M q} \, \Big| \sum_{\{i,j\} \in \bar{\cal S}_q } 
			   [\tilde{\bm{\Theta}} - \bar{\bm \Phi}_k^{-1}]_{ij} \, [{\bm \Lambda}_k]_{ji} \Big| \, , \\
		L_{2k} & = \frac{2}{2 M q} \, \Big| \sum_{\{i,j\} \in \bar{\cal S}_q^c } 
			   [\tilde{\bm{\Theta}} - \bar{\bm \Phi}_k^{-1}]_{ij} \, [{\bm \Lambda}_k]_{ji} \Big|  
\end{align*}
where $\bar{\cal S}_q = {\cal S}_q \cup \{ \{i,j\}: \, i=j \}$ so that $|\bar{\cal S}_q| = s_q+q$.
Using Lemma 5 and $\big| \sum_{\{i,j\} \in \bar{\cal S}_q }  [{\bm \Lambda}_k]_{ij} \big| \le \sqrt{s_q+q} \, \| {\bm \Lambda}_k \|_F$ (by Cauchy-Schwarz inequality), we have 
\begin{align}
 L_{1k} &  \le  \frac{1}{ M q} C_{0q} \sqrt{\frac{\ln(M^{1/\tau} q)}{Kp}} \,
       \Big| \sum_{\{i,j\} \in \bar{\cal S}_q } \,  [{\bm \Lambda}_k]_{ij} \Big|  \nonumber \\
		& \le  \frac{C_{0q}}{ M^{3/2} q} \, r_{qn} \, \| {\bm \Lambda}_k \|_F \, , \label{apeqn630} \\
 L_{2k} &  \le  \frac{C_{0q}}{ M q} \, \sqrt{\frac{\ln(M^{1/\tau} q)}{Kp}} \,
            \| {\bm \Lambda}_{k {\cal S}_q^c}^-\|_1 \, . \label{apeqn631}
\end{align}
Alternatively, as in \cite[Eqn.\ (B.56)]{Tugnait22c}, with $B_2 = \sum_{k=1}^M B_{2k}$,
\begin{align} 
     |B_2| \le & \frac{2}{2Mq} \sum_{i,j=1}^p \sum_{k=1}^M \big| [\tilde{\bm{\Theta}} - \bar{\bm \Phi}_k^{-1}]_{ij} \big|
		   \, \big| [\bm{\Lambda}_k]_{ji} \big| \nonumber \\
	   \le & \frac{C_{0q}}{ M q} \, \sqrt{\frac{\ln(M^{1/\tau} q)}{Kp}} \, 
		 \sum_{i,j=1}^p \sum_{k=1}^M  \big| [\bm{\Lambda}_k]_{ij} \big| \, . \label{apeqn633}
\end{align}
Define $\check{\bm \Lambda} \in \mathbb{R}^{q \times q}$ with $[\check{\bm \Lambda}]_{ij} = \| {\bm \Lambda}^{(ij)} \|_F$ and as in (\ref{eqn232}), ${\bm{\Lambda}}^{(ij)} :=  [ [{\bm{\Lambda}}_1 ]_{ij} \;  \cdots \; [{\bm{\Lambda}}_M ]_{ij}]^\top \in \mathbb{C}^M$. Using $\sum_{k=1}^M  \big| [\bm{\Lambda}_k]_{ij} \big| \le \sqrt{M} \, \| {\bm \Lambda}^{(ij)} \|_F$, we have
\begin{align} 
     |B_2| \le & \frac{C_{0q}}{ \sqrt{M} \, q} \, \sqrt{\frac{\ln(M^{1/\tau} q)}{Kp}} \, 
		 \| \check{\bm \Lambda} \|_1 \, . \label{apeqn634}
\end{align}
Mimicking \cite[Eqns.\ (B.56)-(B.58)]{Tugnait22c}, we have $B_{3k} \ge \alpha \lambda_q (  \| {\bm \Lambda}_{k {\cal S}_q^c}^-\|_1 - \| {\bm \Lambda}_{k {\cal S}_q}^-\|_1 )$ and $B_4 \ge (1-\alpha) \, \sqrt{M} \,  \lambda_q (  \| \check{\bm \Lambda}_{k {\cal S}_q^c}^-\|_1 - \| \check{\bm \Lambda}_{k {\cal S}_q}^-\|_1 )$. With $B_3 = \sum_{k=1}^M B_{3k}$ and using (\ref{apeqn630}) and (\ref{apeqn631}), similar to \cite[Eqns.\ (B.60)]{Tugnait22c}, we have
\begin{align}
  \alpha & B_2 + B_3  \ge -\alpha |B_2| + B_3 \nonumber \\
			& \ge - \alpha \lambda_q \sum_{k=1}^M \| {\bm \Lambda}_{k {\cal S}_q}^-\|_1 
			  - \alpha \frac{C_{0q}}{ M^{3/2} q} \, r_{qn} \, \sum_{k=1}^M \| {\bm \Lambda}_k \|_F  \label{apeqn635}
\end{align}
where we also used the first inequality in (\ref{eqn462}). Using $\| {\bm \Lambda}_{k {\cal S}_q}^- \|_1 \le \sqrt{s_q} \, \| {\bm \Lambda}_k \|_F$,  $\sum_{k=1}^M \| {\bm \Lambda}_k \|_F \le \sqrt{M} \, \|{\bm \Lambda} \|_F$ and  the second inequality in (\ref{eqn462}), we can simplify (\ref{apeqn635}) as
\begin{align}
  \alpha  B_2 + B_3  &  \ge - 2 \alpha  \|{\bm \Lambda} \|_F \, \frac{C_{0q}}{ M q} \, r_{qn} \, . \label{apeqn636}
\end{align}
In a similar manner (see also \cite[Eqns.\ (B.61)]{Tugnait22c}) using (\ref{apeqn634}), we have
\begin{align}
  (1-\alpha)  B_2 + B_4  &  \ge - 2 (1-\alpha)  \|{\bm \Lambda} \|_F \, \frac{C_{0q}}{ M q} \, r_{qn}  \label{apeqn637}
\end{align}
under the upperbound on $\lambda_{qn}$ specified in (\ref{eqn462}). Thus, by (\ref{apeqn610}), (\ref{apeqn622}), (\ref{apeqn636}) and (\ref{apeqn637}), we obtain 
\begin{align} 
	  J({\bm \Lambda})  \ge & \frac{ \| \bm{\Lambda} \|_F^2 \, \beta_{q,\min}^2}{  8 M q } -
		  \frac{2 C_{0q} r_{qn}  \| {\bm \Lambda} \|_F}{M q} \nonumber \\
		  = & 
			 \frac{ \| \bm{\Lambda} \|_F^2 \, \beta_{q,\min}^2}{  8 Mq } \Big( 1- \frac{16}{17} \Big) > 0  \label{apeqn640}
\end{align}
using $\| {\bm \Lambda} \|_F = R_q r_{qn}$ and $R_q = 17 C_{0q}/\beta_{q,\min}^2$. This proves Theorem 2(ii). $\quad \Box$

\section{Proof of Theorem 3} \label{append3}
{\it Proof of Theorem 3(i)}. Since $\| {\bm \Omega}^\diamond \|_F =1$, we have $\bar{\bm \Omega}({\bm \Gamma})/ \|\bar{\bm \Omega}({\bm \Gamma})\|_F = {\bm \Omega}^\diamond$. We have
\begin{align}  
  & \| \hat{\bm \Omega} - {\bm \Omega}^\diamond \|_F = 
	  \Big\| \hat{\bm \Omega}({\bm \Gamma})/ \|\hat{\bm \Omega}({\bm \Gamma})\|_F - 
		  \bar{\bm \Omega}({\bm \Gamma})/ \|\bar{\bm \Omega}({\bm \Gamma})\|_F \Big\|_F  \nonumber \\
	&  =   \Big\| \frac{\hat{\bm \Omega}({\bm \Gamma}) }{ \|\hat{\bm \Omega}({\bm \Gamma})\|_F} - 
	  \frac{\bar{\bm \Omega}({\bm \Gamma})}{ \|\hat{\bm \Omega}({\bm \Gamma})\|_F }    
	 +   \frac{ \bar{\bm \Omega}({\bm \Gamma}) }{ \|\hat{\bm \Omega}({\bm \Gamma})\|_F } -
		  \frac{\bar{\bm \Omega}({\bm \Gamma})}{ \|\bar{\bm \Omega}({\bm \Gamma})\|_F} \Big\|_F  \nonumber \\
	&	 \le \frac{\|\hat{\bm \Omega}({\bm \Gamma}) - \bar{\bm \Omega}({\bm \Gamma}) \|_F}{ \|\hat{\bm \Omega}({\bm \Gamma})\|_F}
	  + \|\bar{\bm \Omega}({\bm \Gamma})\|_F \;
		\Big| \frac{1}{ \|\hat{\bm \Omega}({\bm \Gamma})\|_F} -\frac{1}{ \|\bar{\bm \Omega}({\bm \Gamma})\|_F } \Big|  \nonumber \\
	&	 \le \frac{2 }{ \|\hat{\bm \Omega}({\bm \Gamma})\|_F} \; \|\hat{\bm \Omega}({\bm \Gamma}) - \bar{\bm \Omega}({\bm \Gamma}) \|_F
		\label{apeqn700}
\end{align}
using $\big| \|\bar{\bm \Omega}({\bm \Gamma})\|_F - \|\hat{\bm \Omega}({\bm \Gamma})\|_F \big| \le \|\hat{\bm \Omega}({\bm \Gamma}) - \bar{\bm \Omega}({\bm \Gamma}) \|_F$ (by triangle inequality). Now $\|\hat{\bm \Omega}({\bm \Gamma})\|_F = \|\hat{\bm \Omega}({\bm \Gamma}) -\bar{\bm \Omega}({\bm \Gamma}) + \bar{\bm \Omega}({\bm \Gamma})\|_F \ge \|\bar{\bm \Omega}({\bm \Gamma})\|_F - \|\hat{\bm \Omega}({\bm \Gamma}) - \bar{\bm \Omega}({\bm \Gamma})\|_F$. For $n > N_{2p}$, we have $\|\hat{\bm \Omega}({\bm \Gamma}) - \bar{\bm \Omega}({\bm \Gamma})\|_F \le 0.5 \|\bar{\bm \Omega}({\bm \Gamma})\|_F$, and therefore, $\|\hat{\bm \Omega}({\bm \Gamma})\|_F \ge 0.5 \|\bar{\bm \Omega}({\bm \Gamma})\|_F$. Hence, 
\begin{align}
      \| \hat{\bm \Omega} - {\bm \Omega}^\diamond \|_F 
        &  \le 4 \, \|\hat{\bm \Omega}({\bm \Gamma}) - \bar{\bm \Omega}({\bm \Gamma})\|_F / \|\bar{\bm \Omega}({\bm \Gamma})\|_F \, . \label{apeqn702}
\end{align} 
We now characterize $\|\bar{\bm \Omega}({\bm \Gamma})\|_F$. We have
\begin{align*}  
    A & = \big| \sum_{k=1}^M \big( \mbox{tr}( \bar{\bm S}^\diamond_k {\bm \Phi}_k)
	  + \mbox{tr}( \bar{\bm S}^\diamond_k {\bm \Phi}_k)^\ast \big) \big| 
		  \le 2 \sum_{k=1}^M \big| \mbox{tr}( \bar{\bm S}^\diamond_k {\bm \Phi}_k) \big | \\ 
		&  \le 2 \sum_{k=1}^M \| \bar{\bm S}^\diamond_k \|_F \, \| {\bm \Phi}_k \|_F
		   \le 2 \sqrt{q}\, \beta_{q,\max} \, \sum_{k=1}^M   \| {\bm \Phi}_k \|_F \, . 
\end{align*}
Since $\sum_{k=1}^M \| {\bm \Phi}_k \|_F \le \sum_{k=1}^M \| {\bm \Phi}_k - {\bm \Phi}_k^\diamond \|_F + \sum_{k=1}^M \| {\bm \Phi}_k^\diamond \|_F
 \le \sqrt{M} \, \gamma_q +  \sqrt{q} \, M /\beta_{q,\min}$, we have $A  \le 0.2 \sqrt{qM} + 2 qM \beta_{q,\max} /\beta_{q,\min} \le 2qM (1+\beta_{q,\max} /\beta_{q,\min})$. By (\ref{aeqn254}) and the fact $\| {\bm \Omega}^\diamond \|_F =1$, we infer $\|\bar{\bm \Omega}({\bm \Gamma})\|_F \ge \beta_{q,\min} /(\beta_{q,\max}+\beta_{q,\min} ) = 1/ \gamma_r $, which combined with (\ref{apeqn702}) and (\ref{eqn455}) yields  (\ref{eqn467}). 
$\quad \Box$ \\
{\it Proof of Theorem 3(ii)}. For $n > N_{3p}$, $\hat{\bm \Omega} \in {\cal B}({\bm \Omega}^\diamond)$ (cf.\ Theorem 3(i)), and $C_{2p} r_{pn} \le (p/2)$ w.h.p. We have
\begin{align*} 
 & \| \hat{\bm \Gamma}(\hat{\bm \Omega}) - {\bm \Gamma}^\diamond \|_F  
	   \le \| \hat{\bm \Gamma}(\hat{\bm \Omega}) - \bar{\bm \Gamma}(\hat{\bm \Omega}) \|_F
		  + \| \bar{\bm \Gamma}(\hat{\bm \Omega}) - {\bm \Gamma}^\diamond \|_F  
\end{align*}
where Theorem 2(ii) applies to $\| \hat{\bm \Gamma}(\hat{\bm \Omega}) - \bar{\bm \Gamma}(\hat{\bm \Omega}) \|_F$. By (\ref{aeqn256}),
\begin{align} 
  \bar{\bm \Gamma}(\hat{\bm \Omega}) & - {\bm \Gamma}^\diamond 
	   = \big( \frac{p}{\mbox{tr}( {\bm \Sigma}^\diamond \hat{\bm \Omega} )} -1 \big)  {\bm \Gamma}^\diamond \, . \label{apeqn706}
\end{align} 
As in the proof of Theorem 2(ii) (following (\ref{apeqn620})), we have $\mbox{tr}( {\bm \Sigma}^\diamond \hat{\bm \Omega} ) = \mbox{tr}( {\bm \Sigma}^\diamond (\hat{\bm \Omega} -{\bm \Omega}^\diamond + {\bm \Omega}^\diamond)) = \mbox{tr}( {\bm \Sigma}^\diamond (\hat{\bm \Omega} -{\bm \Omega}^\diamond) ) +p$ and $|\mbox{tr}( {\bm \Sigma}^\diamond (\hat{\bm \Omega} -{\bm \Omega}^\diamond) )| \le \|{\bm \Sigma}^\diamond\|_F \,  \|\hat {\bm \Omega} -{\bm \Omega}^\diamond \|_F \le C_{2p} r_{pn}$ (using (\ref{eqn467})). Therefore, $p-C_{2p}r_{pn} \le \mbox{tr}( {\bm \Sigma}^\diamond \hat{\bm \Omega} )  \le p+C_{2p}r_{pn}$ and $ |p-\mbox{tr}( {\bm \Sigma}^\diamond \hat{\bm \Omega} )|  \le C_{2p}r_{pn}$. Since $0 < C_{2p} r_{pn} \le (p/2)$ w.h.p., $|\mbox{tr}( {\bm \Sigma}^\diamond \hat{\bm \Omega} )|^{-1} \le 2/p$. Thus we have $\| \bar{\bm \Gamma}(\hat{\bm \Omega}) - {\bm \Gamma}^\diamond \|_F \le C_{2q} r_{pn}$, which yields (\ref{eqn469}). The given probability bound is the result of the bounds in Theorem 2 (both (\ref{eqn455}) and (\ref{eqn465}) must hold) and an application of the union bound. $\quad \Box$

\bibliographystyle{unsrt} %

\end{document}